\documentclass[10pt, letter, onecolumn]{arxiv}

\usepackage[bibstyle=nature,citestyle=numeric-comp,%
            natbib=true,backend=biber,maxbibnames=99,%
            giveninits=false,sorting=none]{biblatex}
\usepackage{nameref}
\usepackage{varioref}
\usepackage[noabbrev,capitalize]{cleveref}
\usepackage{etoc}

\graphicspath{{figures/}}

\usepackage{graphicx}
\usepackage{pifont}
\usepackage{amssymb}
\usepackage{multirow}
\usepackage{array}
\usepackage{dblfloatfix}
\usepackage{titlesec}
\usepackage{graphicx}
\newlength{\cboxlength}
\usepackage{amsmath,xparse,mleftright}
\NewDocumentCommand{\up}{som}{%
  \IfBooleanTF{#1}
    {\upext{#3}}
    {#3\IfNoValueTF{#2}{\mathord}{#2}\uparrow}%
}
\NewDocumentCommand{\upext}{m}{%
  \mleft.\kern-\nulldelimiterspace#1\mright\uparrow
}

\usepackage{xcolor} % For text color
\usepackage{soul} % For highlighting
\ifdefined\red
    \renewcommand{\red}[1]{\textcolor{red}{#1}}
\else
    \newcommand{\red}[1]{\textcolor{red}{#1}}
\fi
\ifdefined\blue
    \renewcommand{\blue}[1]{\textcolor{blue}{#1}}
\else
    \newcommand{\blue}[1]{\textcolor{blue}{#1}}
\fi
\newcommand{\hhll}[1]{#1} % highlights off

\definecolor{d1_r1_q1}{RGB}{166,166,255} 
\definecolor{d1_r1_q2}{RGB}{212,212,255} 
\definecolor{d1_r1_q3}{RGB}{225,225,255} 
\definecolor{d1_r2_q1}{RGB}{164,164,255} 
\definecolor{d1_r2_q2}{RGB}{192,192,255} 
\definecolor{d1_r2_q3}{RGB}{255,248,248} 
\definecolor{d1_r3_q1}{RGB}{235,235,255} 
\definecolor{d1_r3_q2}{RGB}{240,240,255} 
\definecolor{d1_r3_q3}{RGB}{255,212,212} 
\definecolor{d1_r4_q1}{RGB}{136,136,255} 
\definecolor{d1_r4_q2}{RGB}{182,182,255} 
\definecolor{d1_r4_q3}{RGB}{225,225,255} 
\definecolor{d1_r5_q1}{RGB}{220,220,255} 
\definecolor{d1_r5_q2}{RGB}{240,240,255} 
\definecolor{d1_r5_q3}{RGB}{255,240,240} 
\definecolor{d1_ra_q1}{RGB}{184,184,255} 
\definecolor{d1_ra_q2}{RGB}{212,212,255} 
\definecolor{d1_ra_q3}{RGB}{255,255,255} 
\definecolor{d2_r1_q1}{RGB}{207,207,255} 
\definecolor{d2_r1_q2}{RGB}{235,235,255} 
\definecolor{d2_r1_q3}{RGB}{248,248,255} 
\definecolor{d2_r2_q1}{RGB}{230,230,255} 
\definecolor{d2_r2_q2}{RGB}{255,253,253} 
\definecolor{d2_r2_q3}{RGB}{253,253,255} 
\definecolor{d2_r3_q1}{RGB}{197,197,255} 
\definecolor{d2_r3_q2}{RGB}{204,204,255} 
\definecolor{d2_r3_q3}{RGB}{199,199,255} 
\definecolor{d2_r4_q1}{RGB}{207,207,255} 
\definecolor{d2_r4_q2}{RGB}{255,255,255} 
\definecolor{d2_r4_q3}{RGB}{255,255,255} 
\definecolor{d2_r5_q1}{RGB}{233,233,255} 
\definecolor{d2_r5_q2}{RGB}{245,245,255} 
\definecolor{d2_r5_q3}{RGB}{245,245,255} 
\definecolor{d2_ra_q1}{RGB}{215,215,255} 
\definecolor{d2_ra_q2}{RGB}{240,240,255} 
\definecolor{d2_ra_q3}{RGB}{240,240,255} 
\definecolor{d3_r1_q1}{RGB}{169,169,255} 
\definecolor{d3_r1_q2}{RGB}{243,243,255} 
\definecolor{d3_r1_q3}{RGB}{253,253,255} 
\definecolor{d3_r2_q1}{RGB}{197,197,255} 
\definecolor{d3_r2_q2}{RGB}{240,240,255} 
\definecolor{d3_r2_q3}{RGB}{245,245,255} 
\definecolor{d3_r3_q1}{RGB}{187,187,255} 
\definecolor{d3_r3_q2}{RGB}{230,230,255} 
\definecolor{d3_r3_q3}{RGB}{233,233,255} 
\definecolor{d3_r4_q1}{RGB}{192,192,255} 
\definecolor{d3_r4_q2}{RGB}{243,243,255} 
\definecolor{d3_r4_q3}{RGB}{212,212,255} 
\definecolor{d3_r5_q1}{RGB}{204,204,255} 
\definecolor{d3_r5_q2}{RGB}{255,235,235} 
\definecolor{d3_r5_q3}{RGB}{255,253,253} 
\definecolor{d3_ra_q1}{RGB}{189,189,255} 
\definecolor{d3_ra_q2}{RGB}{245,245,255} 
\definecolor{d3_ra_q3}{RGB}{240,240,255} 
\definecolor{d1_ra_q1}{RGB}{184,184,255} 
\definecolor{d1_ra_q2}{RGB}{212,212,255} 
\definecolor{d1_ra_q3}{RGB}{255,255,255} 
\definecolor{d2_ra_q1}{RGB}{215,215,255} 
\definecolor{d2_ra_q2}{RGB}{240,240,255} 
\definecolor{d2_ra_q3}{RGB}{240,240,255} 
\definecolor{d3_ra_q1}{RGB}{189,189,255} 
\definecolor{d3_ra_q2}{RGB}{245,245,255} 
\definecolor{d3_ra_q3}{RGB}{240,240,255} 
\definecolor{d99_r99_q1}{RGB}{197,197,255} 
\definecolor{d99_r99_q2}{RGB}{235,235,255} 
\definecolor{d99_r99_q3}{RGB}{245,245,255} 

\makeatletter
\renewcommand\AB@authnote[1]{}
\renewcommand\AB@affilnote[1]{}
\makeatother

\usepackage{anyfontsize} % allows for precise control over title font size

\title{{\fontsize{17.5pt}{16.5pt}\selectfont Adapted Large Language Models Can Outperform Medical Experts in Clinical Text Summarization}} % first value is the font size, and the second value is the line spacing 

\author[]{Dave Van Veen$^{\dagger}$, Cara Van Uden, Louis Blankemeier, Jean-Benoit Delbrouck, Asad Aali$^{\ddagger}$, \\ Christian Bluethgen, Anuj Pareek, Malgorzata Polacin, Eduardo Pontes Reis, \\ Anna Seehofnerová, Nidhi Rohatgi, Poonam Hosamani, William Collins, Neera Ahuja, \\Curtis P. Langlotz, Jason Hom, Sergios Gatidis, John Pauly, Akshay S. Chaudhari}

\affil{Stanford University}

\renewcommand{\correspondingauthor}[1]{
                                    Published in Nature Medicine
                                    \; \;
                                    $\dagger$~Corresponding author: vanveen@stanford.edu \; \;
                                    $\ddagger$~The University of Texas at Austin
                                    }

\begin{document}
\begin{refsection}

\begin{abstract}

Analyzing vast textual data and summarizing key information from electronic health records imposes a substantial burden on how clinicians allocate their time. Although large language models (LLMs) have shown promise in natural language processing (NLP), their effectiveness on a diverse range of clinical summarization tasks remains unproven. In this study, we apply adaptation methods to eight LLMs, spanning four distinct clinical summarization tasks: radiology reports, patient questions, progress notes, and doctor-patient dialogue. Quantitative assessments with syntactic, semantic, and conceptual NLP metrics reveal trade-offs between models and adaptation methods. A clinical reader study with ten physicians evaluates summary completeness, correctness, and conciseness; in a majority of cases, summaries from our best adapted LLMs are either equivalent (45\%) or superior (36\%) compared to summaries from medical experts. The ensuing safety analysis highlights challenges faced by both LLMs and medical experts, as we connect errors to potential medical harm and categorize types of fabricated information. Our research provides evidence of LLMs outperforming medical experts in clinical text summarization across multiple tasks. This suggests that integrating LLMs into clinical workflows could alleviate documentation burden, allowing clinicians to focus more on patient care.

\end{abstract}

\maketitle

% ------------ SECTIONS ---------------------

\vspace{10mm}
\section{Introduction}

Documentation plays an indispensable role in healthcare practice. Currently, clinicians spend significant time summarizing vast amounts of textual information---whether it be compiling diagnostic reports, writing progress notes, or synthesizing a patient’s treatment history across different specialists~\cite{golob2016painful, arndt2017tethered, fleming2023medalign}. Even for experienced physicians with a high level of expertise, this intricate task naturally introduces the possibility for errors, which can be detrimental in healthcare where precision is paramount~\cite{yackel2010unintended, bowman2013impact, gershanik2011critical}.

The widespread adoption of electronic health records has expanded clinical documentation workload, directly contributing to increasing stress and clinician burnout~\cite{gesner2019burden, ratwani2018usability, ehrenfeld2018technology}. Recent data indicates that physicians can
expend up to two hours on documentation for each hour of patient interaction~\cite{sinsky2016allocation}. Meanwhile, documentation responsibilities for nurses can consume up to 60\% of their time and account for significant work stress~\cite{khamisa2013burnout, duffy2010point, chang2016nurses}. These tasks divert attention from direct patient care, leading to worse outcomes for patients and decreased job satisfaction for clinicians~\cite{arndt2017tethered, shanafelt2016relationship, robinson2018novel, toussaint2020design}.

In recent years, large language models (LLMs) have gained remarkable traction, leading to widespread adoption of models such as ChatGPT~\cite{brown2020language}, which excel at information retrieval, nuanced understanding, and text generation~\cite{zhao2023survey, bubeck2023sparks}. Although LLM benchmarks for general natural language processing (NLP) tasks exist~\cite{liang2022holistic, zheng2023judging}, they do not evaluate performance on relevant clinical tasks. Addressing this limitation presents an opportunity to accelerate the process of clinical text summarization, hence alleviating documentation burden and improving patient care.

Crucially, machine-generated summaries must be non-inferior to that of seasoned clinicians—especially when used to support sensitive clinical decision-making. Previous work has demonstrated potential across clinical NLP tasks~\cite{wornow2023shaky, thirunavukarasu2023large}, adapting to the medical domain by either training a new model~\cite{singhal2022large, tu2023towards}, fine-tuning an existing model~\cite{toma2023clinical, van2023radadapt}, or supplying domain-specific examples in the model prompt~\cite{mathur2023summqa, van2023radadapt}. However, adapting LLMs to summarize a diverse set of clinical tasks has not been thoroughly explored, nor has non-inferiority to medical experts been achieved. With the overarching objective of bringing LLMs closer to clinical readiness, we make the following contributions:

\begin{itemize}
    \item We implement adaptation methods across eight open-source and proprietary LLMs for four distinct summarization tasks comprising six datasets. The subsequent evaluation via NLP metrics provides a comprehensive assessment of contemporary LLMs for clinical text summarization.
    \item Our exploration delves into a myriad of trade-offs concerning different models and adaptation methods, shedding light on scenarios where advancements in model size, novelty, or domain specificity do not necessarily translate to superior performance.
    \item Through a clinical reader study with ten physicians, we demonstrate that LLM summaries can surpass medical expert summaries in terms of completeness, correctness, and conciseness.
    \item Our safety analysis of examples, potential medical harm, and fabricated information reveals insights into the challenges faced by both models and medical experts.
    \item We identify which NLP metrics most correlate with reader preferences.
    
\end{itemize}

Our study demonstrates that adapting LLMs can outperform medical experts for clinical text summarization across the diverse range of documents we evaluate. This suggests that incorporating LLM-generated candidate summaries could reduce documentation load, potentially leading to decreased clinician strain and improved patient care.

\section{Related Work}
\begin{figure*}[t]
    \centering
    \includegraphics[width = 1\textwidth]{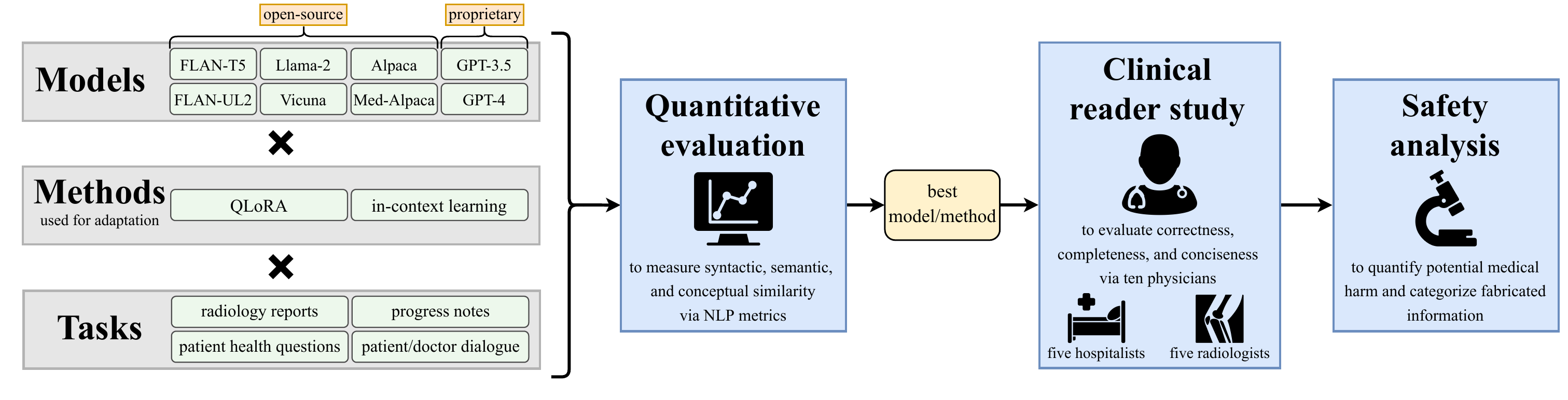}
    \caption{
    \textbf{Framework overview.} First, we quantitatively evaluate each valid combination ($\times$) of LLM and adaptation method across four distinct summarization tasks comprising six datasets. We then conduct a clinical reader study in which ten physicians compare summaries of the best model/method against those of a medical expert. Lastly, we perform a safety analysis to quantify potential medical harm and to categorize types of fabricated information.
    }
    \label{fig:overview}
\end{figure*}

Large language models (LLMs) have demonstrated astounding performance, propelled by both the transformer architecture~\cite{vaswani2017attention} and increasing scales of data and compute, resulting in widespread adoption of models such as ChatGPT~\cite{brown2020language}.
Although several of the more expansive models, such as GPT-4~\cite{openai2023gpt4} and PaLM~\cite{chowdhery2022palm}, remain proprietary and provide access via ``black-box'' interfaces, there has been a pronounced shift towards open-sourced alternatives such as Llama-2~\cite{touvron2023llama}. These open-source models grant researchers direct access to model weights for customization.

Popular transformer models such as BERT~\cite{devlin2018bert} and GPT-2~\cite{radford2019language} established the paradigm of self-supervised pretraining on large amounts of general data and then adapting to a particular domain or task by tuning on specific data. One approach is customizing model weights via instruction tuning, a process where language models are trained to generate human-aligned responses given specific instructions~\cite{wei2021finetuned}. Examples of clinical instruction-tuned models include Med-PALM~\cite{singhal2022large} for medical question-answering or Radiology-GPT~\cite{liu2023radiology} for radiology tasks. To enable domain adaptation with limited computational resources, prefix tuning~\cite{li2021prefix} and low-rank adaptation (LoRA)~\cite{hu2021lora} have emerged as effective methods that require tuning less than 1\% of total parameters over a small training set. LoRA has been shown to work well for medical question-answering~\cite{toma2023clinical} and summarizing radiology reports~\cite{van2023radadapt}. Another adaptation method, requiring no parameter tuning, is in-context learning: supplying the LLM with task-specific examples in the prompt~\cite{lampinen2022can}. Because in-context learning does not alter model weights, it can be performed with black-box model access using only a few training examples~\cite{lampinen2022can}.

Recent work has adapted LLMs for various medical tasks, demonstrating great potential for medical language understanding and generation~\cite{wornow2023shaky, thirunavukarasu2023large, tu2023towards, liu2023leveraging}.
Specifically, a broad spectrum of methodologies has been applied to clinical text for specific summarization tasks. One such task is the summarization of radiology reports, which aims to consolidate detailed findings from radiological studies into significant observations and conclusions drawn by the radiologist~\cite{kahn2009toward}. LLMs have shown promise on this task~\cite{van2023radadapt} and other tasks such as summarizing daily progress notes into a concise ``problem list'' of medical diagnoses~\cite{gao2023overview}. Lastly, there has been significant work on summarizing extended conversations between a doctor and patient into patient visit summaries~\cite{abacha2023overview, yim2023aci, mathur2023summqa}.

While the aforementioned contributions incorporate methods to adapt language models, they often include only a small subset of potential approaches and models, and/or they predominantly rely on evaluation via standard NLP metrics. Given the critical nature of medical tasks, demonstrating clinical readiness requires including human experts in the evaluation process. To address this, there have been recent releases of expert evaluations for instruction following~\cite{fleming2023medalign} and radiology report generation~\cite{yu2023radiology}. Other work employs human experts to evaluate synthesized Cochrane review abstracts, demonstrating that NLP metrics are not sufficient to measure summary quality~\cite{tang2023evaluating}. With this in mind, we 
extend our comprehensive evaluation of methods and LLMs beyond NLP metrics to incorporate a clinical reader study across multiple summarization tasks. Our results demonstrate across many tasks that LLM summaries are comparable to–––and often surpass–––those created by human experts.
\section{Approach}
\label{sec:methods}

\subsection{Large language models}
\setlength{\tabcolsep}{6pt} % spacing b/w cols, default 6pt

\begin{table*}[b!]
\caption{We quantitatively evaluate eight models, including state-of-the-art sequence-to-sequence and autoregressive models. Unless specified, models are open-source (vs. proprietary).
}
\vspace{-2mm}
\begin{center}
\begin{tabular}{l | c c | c | c c}
 \textbf{Model} & \textbf{Context} & \textbf{Parameters} & \textbf{Proprietary?} & \textbf{Seq2seq} &\textbf{Autoreg.} \\
\hline
FLAN-T5 & 512 & 2.7B & - & \ding{52} & - \\ 
FLAN-UL2 & 2,048 & 20B & - & \ding{52} & - \\ 
Alpaca & 2,048 & 7B & - & - & \ding{52} \\
Med-Alpaca & 2,048 & 7B & - & - & \ding{52} \\
Vicuna & 2,048 & 7B & - & - & \ding{52} \\
Llama-2 & 4,096 & 7B, 13B & - & - & \ding{52} \\
GPT-3.5 & 16,384 & 175B & \ding{52} & - & \ding{52} \\
GPT-4 & 32,768* & unknown & \ding{52} & - & \ding{52} \\
 \multicolumn{6}{l}{\footnotesize \hhll{*The context length of GPT-4 has since been increased to 128,000.}} \\
\end{tabular}
\end{center}
\label{tab:models}
\end{table*}

We investigate a diverse collection of transformer-based LLMs for clinical summarization tasks. This includes two broad approaches to language generation: sequence-to-sequence (seq2seq) models and autoregressive models. Seq2seq models use an encoder-decoder architecture to map the input text to a generated output, often requiring paired datasets for training. These models have shown strong performance in machine translation~\cite{chen2018best} and summarization~\cite{shi2021neural}. In contrast, the autoregressive models typically only use a decoder. They generate tokens sequentially---where each new token is conditioned on previous tokens---thus efficiently capturing context and long-range dependencies. Autoregressive models are typically trained with unpaired data, and they are particularly useful for various NLP tasks such as text generation, question-answering, and dialogue interactions~\cite{vicuna2023, brown2020language}.

We include prominent seq2seq models due to their strong summarization performance~\cite{shi2021neural} and autoregressive models due to their state-of-the-art performance across general NLP tasks~\cite{zheng2023judging}. As shown in Table~\ref{tab:models}, our choice of models varies widely with respect to number of parameters (2.7 billion to 175 billion) and context length (512 to 32,768), i.e.~the maximum number of input tokens a model can process. We organize our models into three categories:

\textbf{Open-source seq2seq models}. The original T5 ``text-to-text transfer transformer'' model~\cite{raffel2020t5} demonstrated excellent performance in transfer learning using the seq2seq architecture. A derivative model, FLAN-T5~\cite{chung2022flant5,longpre2023flan}, improved performance via instruction prompt tuning. This T5 model family has proven effective for various clinical NLP tasks~\cite{lehman2023we, van2023radadapt}. The FLAN-UL2 model~\cite{tay2022ul2,chung2022scaling} was introduced recently, which featured an increased context length (four-fold that of FLAN-T5) and a modified pre-training procedure called unified language learning (UL2).

\textbf{Open-source autoregressive models}. The Llama family of LLMs~\cite{touvron2023llama} has enabled the proliferation of open-source instruction-tuned models that deliver comparable performance to GPT-3~\cite{brown2020language} on many benchmarks despite their smaller sizes. Descendants of this original model have taken additional fine-tuning approaches, such as fine-tuning via instruction following (Alpaca~\cite{alpaca}), medical Q\&A data (Med-Alpaca~\cite{han2023medalpaca}), user-shared conversations (Vicuna~\cite{vicuna2023}), and reinforcement learning from human feedback (Llama-2~\cite{touvron2023llama}). Llama-2 allows for two-fold longer context lengths (4,096) relative to the aforementioned open-source autoregressive models.

\textbf{Proprietary autoregressive models}. We include GPT-3.5~\cite{openai2022chatgpt} and GPT-4~\cite{openai2023gpt4}, the latter of which has been regarded as state-of-the-art on general NLP tasks~\cite{zheng2023judging}\hhll{ and has demonstrated strong performance on biomedical NLP tasks such as medical exams}~\cite{lim2023benchmarking, rosol2023evaluation, brin2023comparing}. Both models offer significantly higher context length (16,384 and 32,768) than open-source models. \hhll{We note that since sharing our work, GPT-4's context length has been increased to 128,000.} 

\subsection{Adaptation methods}\label{sec:methods_adaptation_techniques}

We consider two proven techniques for adapting pre-trained general-purpose LLMs to domain-specific clinical summarization tasks. To demonstrate the benefit of adaptation methods, we also include the baseline zero-shot prompting, i.e.~$m=0$ in-context examples.

\textbf{In-context learning (ICL)}. ICL is a lightweight adaptation method that requires no altering of model weights; instead, one includes a handful of in-context examples directly within the model prompt~\cite{lampinen2022can}. This simple approach provides the model with context, enhancing LLM performance for a particular task or domain~\cite{mathur2023summqa, van2023radadapt}. We implement this by choosing, for each sample in our test set, the $m$ nearest neighbors training samples in the embedding space of the PubMedBERT model~\cite{deka2022evidence}. Note that choosing ``relevant'' in-context examples has been shown to outperform choosing examples at random~\cite{nie2022improving}. For a given model and dataset, we use $m=2^x$ examples, where $x \in \{0, 1, 2, 3, ..., M\}$ for $M$ such that no more than $1\%$ of the $s=250$ samples are excluded due to prompts exceeding the model's context length. Hence each model's context length limits the allowable number of in-context examples.

\textbf{Quantized low-rank adaptation (QLoRA)}. Low-rank adaptation (LoRA)~\cite{hu2021lora} has emerged as an effective, lightweight approach for fine-tuning LLMs by altering a small subset of model weights---often $< 0.1\%$~\cite{van2023radadapt}. LoRA inserts trainable matrices into the attention layers; then, using a training set of samples, this method performs gradient descent on the inserted matrices while keeping the original model weights frozen.  
Compared to training model weights from scratch, LoRA is much more efficient with respect to both computational requirements and the volume of training data required. Recently, QLoRA~\cite{dettmers2023qlora} has been introduced as a more memory-efficient variant of LoRA, employing 4-bit quantization to enable the fine-tuning of larger LLMs given the same hardware constraints. This quantization negligibly impacts performance~\cite{dettmers2023qlora}; as such, we use QLoRA for all model training. Note that QLoRA cannot be used to fine-tune proprietary models on our consumer hardware, as their model weights are not publicly available. \hhll{Fine-tuning of GPT-3.5 via API was made available after our internal model cutoff date of July 31st, 2023}~\cite{openai-gpt3-5-2023}.

\subsection{Data}\label{sec:data-section}
\setlength{\tabcolsep}{6pt} % spacing b/w cols, default 6pt
\definecolor{lightgrray}{gray}{0.9}  % Defining a light gray color

\begin{table*}[b!]
\caption{
\textbf{\underline{\smash{Top}}:} Description of six open-source datasets with a wide range of token length and lexical variance, i.e.~$\frac{\text{number of unique words}}{\text{number of total words}}$.
\textbf{\underline{\smash{Bottom}}:} Instructions for each of the four summarization tasks. See Figure~\ref{fig:prompt-anatomy-engr} for the full prompt.
}
\vspace{-2mm}
\begin{center}
\begin{tabular}{|l l | c c c c|}
\hline
\multicolumn{6}{|c|}{\cellcolor{lightgrray} \textbf{Dataset descriptions}} \\
\hline
& & \textbf{Number} & \multicolumn{2}{c}{\textbf{Avg.~number of tokens}} & \textbf{Lexical} \\
\textbf{Dataset} & \textbf{Task} & \textbf{of samples} & Input & Target  & \textbf{variance} \\
\hline
Open-i & Radiology reports  & 3.4K & 52 $\pm$ 22 & 14 $\pm$ 12 & 0.11 \\
MIMIC-CXR & Radiology reports & 128K & 75 $\pm$ 31 & 22 $\pm$ 17 & 0.08 \\
MIMIC-III & Radiology reports & 67K & 160 $\pm$ 83 & 61 $\pm$ 45 & 0.09 \\
MeQSum & Patient questions & 1.2K & 83 $\pm$ 67 & 14 $\pm$ 6 & 0.21 \\
ProbSum & Progress notes & 755 & 1,013 $\pm$ 299 & 23 $\pm$ 16 & 0.15 \\
ACI-Bench & Dialogue & 126 & 1,512 $\pm$ 467 & 211 $\pm$ 98 & 0.04 \\
\hline
\multicolumn{6}{|c|}{ } \\
\hline
\multicolumn{6}{|c|}{\cellcolor{lightgrray} \textbf{Task Instructions}} \\
\hline
\multicolumn{1}{|l|}{\textbf{Task}} & \multicolumn{5}{|l|}{\textbf{Instruction}} \\
\hline
\rule{0pt}{0.5cm} % add space after hline. requires negative hspace on next line
\hspace{-1.25mm}\multirow{2}{*}{Radiology reports} & \multicolumn{5}{|l|}{``Summarize the radiology report findings} \\
& \multicolumn{5}{|l|}{into an impression with minimal text.''} \\[0.15cm]
\multirow{2}{*}{Patient questions} & \multicolumn{5}{|l|}{``Summarize the patient health query} \\
& \multicolumn{5}{|l|}{into one question of 15 words or less.''} \\[0.15cm]
\multirow{2}{*}{Progress notes} & \multicolumn{5}{|l|}{``Based on the progress note, generate a list of 3-7 prob-} \\
& \multicolumn{5}{|l|}{lems (a few words each) ranked in order of importance.''} \\[0.15cm]
\multirow{2}{*}{Dialogue} & \multicolumn{5}{|l|}{``Summarize the patient/doctor dialogue} \\
& \multicolumn{5}{|l|}{into an assessment and plan.''} \\[0.15cm]
\hline

\end{tabular}
\end{center}
\label{tab:datasets}
\end{table*}

To robustly evaluate LLM performance on clinical text summarization, we choose four distinct summarization tasks, comprising six open-source datasets. As depicted in Table~\ref{tab:datasets}, each dataset contains a varying number of samples, token lengths, and lexical variance. Lexical variance is calculated as
$\frac{\text{number of unique words}}{\text{number of total words}}$
across the entire dataset; hence a higher ratio indicates less repetition and more lexical diversity. We describe each task and dataset below. For task examples, please see Figures~\ref{fig:clin-summ-example-iii},~\ref{fig:clin-summ-example-chq},~\ref{fig:clin-summ-example-pls}, and~\ref{fig:clin-summ-example-d2n}.

\textbf{Radiology reports} Radiology report summarization takes as input the findings section of a radiology study containing detailed exam analysis and results. The goal is to summarize these findings into an impression section, which concisely captures the most salient, actionable information from the study. We consider three datasets for this task, where both reports and findings were created by attending physicians as part of routine clinical care. \underline{\smash{Open-i}}~\cite{demner2016preparing} contains de-identified narrative chest x-ray reports from the Indiana Network for Patient Care 10 database. From the initial set of 4K studies,~\citet{demner2016preparing} selected a final set of 3.4K reports based on the quality of imaging views and diagnostic content. \underline{\smash{MIMIC-CXR}}~\cite{johnson2019mimiccxr} contains chest x-ray studies accompanied by free-text radiology reports acquired at the Beth Israel Deaconess Medical Center between 2011 and 2016. For this study, we use a dataset of 128K reports~\cite{chen-etal-2023-toward} preprocessed by RadSum23 at BioNLP 2023~\cite{DelbrouckRadSum23,demner202322nd}. \underline{\smash{MIMIC-III}}~\cite{johnson2020mimic} contains 67K radiology reports spanning seven anatomies (head, abdomen, chest, spine, neck, sinus, and pelvis) and two modalities: magnetic resonance imaging (MRI) and computed tomography (CT). This dataset originated from patient stays in critical care units of the Beth Israel Deaconess Medical Center between 2001 and 2012. For this study, we utilize a preprocessed version via RadSum23~\cite{DelbrouckRadSum23,demner202322nd}. Compared to x-rays, MRIs and CT scans capture more information at a higher resolution. This usually leads to longer reports (Table~\ref{tab:datasets}), rendering MIMIC-III a more challenging summarization dataset than Open-i or MIMIC-CXR.

\textbf{Patient questions} Question summarization consists of generating a condensed question expressing the minimum information required to find correct answers to the original question~\cite{MeQSum}. For this task, we employ the MeQSum dataset~\cite{MeQSum}. MeQSum contains (1) patient health questions of varying verbosity and coherence selected from messages sent to the U.S. National Library of Medicine (2) corresponding condensed questions created by three medical experts such that the summary allows retrieving complete, correct answers to the original question without the potential for further condensation. These condensed questions were then validated by \hhll{a medical doctor} and verified to have high inter-annotator agreement. Due to the wide variety of these questions, MeQSum exhibits the highest lexical variance of our datasets (Table~\ref{tab:datasets}).

\textbf{Progress notes} The goal of this task is to generate a ``problem list,'' or condensed list of diagnoses and medical problems using the provider's progress notes during hospitalization. For this task, we employ the ProbSum dataset~\cite{gao2023overview}. This dataset\hhll{, generated by attending internal medicine physicians during the course of routine clinical practice,} was extracted from the MIMIC-III database of de-identified hospital intensive care unit (ICU) admissions. ProbSum contains (1) progress notes averaging $> 1,000$ tokens and substantial presence of unlabeled numerical data, e.g.~dates and test results, and (2) corresponding problem lists created by attending medical experts in the ICU. We utilize a version shared by the BioNLP Problem List Summarization Shared Task~\cite{gao2023overview,gao2023bionlp,demner202322nd} and PhysioNet~\cite{PhysioNet}.

\textbf{Dialogue} The goal of this task is to summarize a doctor-patient conversation into an ``assessment and plan'' paragraph. For this task, we employ the ACI-Bench dataset~\cite{yim2023aci, abacha2023overview, MEDIQA-Sum2023}, which contains (1) 207 doctor-patient conversations and (2) corresponding patient visit notes, which were first generated by a seq2seq model and subsequently corrected and validated by expert medical scribes and physicians. Since ACI-Bench's visit notes include a heterogeneous collection of section headers, we choose 126 samples containing an ``assessment and plan'' section for our analysis. Per Table~\ref{tab:datasets}, this task entailed the largest token count across our six datasets for both the input (dialogue) and target (assessment).

\hhll{As we are not the first to employ these datasets, Table A2 contains quantitative metric scores from other works}~\cite{ma2023impressiongpt, van2023radadapt, tu2023towards, wei2023medical, manakul2023cued, yim2023aci}\hhll{ who developed methods specific to each individual summarization task.}
\section{Experiments}
\label{sec:experiments}

This section contains experimental details and study design for our evaluation framework, as depicted in Figure~\ref{fig:overview}.

~\subsection{Quantitative Evaluation}\label{sec:quant_eval}

Building upon the descriptions of models, methods, and tasks in Section~\ref{sec:methods}, we now specify experimental details such as model prompts, data preparation, software implementation, and NLP metrics used for quantitative evaluation. 

~\subsubsection{Model prompts and temperature}\label{sec:prompt}
\begin{figure*}[h!]
    \centering
    \includegraphics[width = \textwidth]{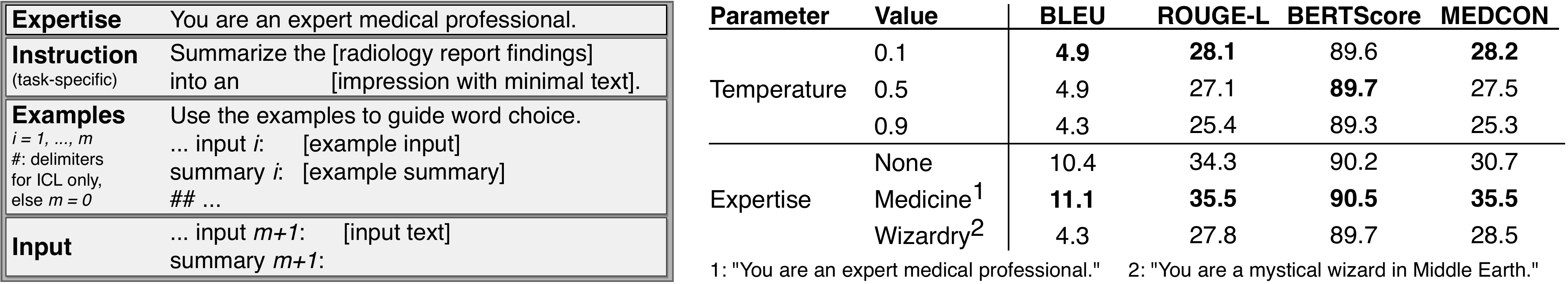}
    \caption{
    \textbf{\underline{\smash{Left}}: Prompt anatomy.}
    Each summarization task uses a slightly different instruction (Table~\ref{tab:datasets}). 
    \textbf{\underline{\smash{Right}}: Effect of model temperature and expertise.}
    We generally find better performance when (1) using lower temperature, i.e. generating less random output, as summarization tasks benefit more from truthfulness than creativity (2) assigning the model clinical expertise in the prompt. Output generated via GPT-3.5 on the Open-i radiology report dataset.
    }
    \label{fig:prompt-anatomy-engr}
\end{figure*}

As shown in Figure~\ref{fig:prompt-anatomy-engr}, we structure prompts by following best practices~\cite{Saravia_Prompt_Engineering_Guide_2022, openai2023bestpractices} and evaluating 1-2 options for model expertise and task-specific instructions (Table~\ref{tab:datasets}). We note the importance of specifying desired length in the instruction, e.g. ``one question of 15 words or less'' for summarizing patient questions. Without this specification, the model might generate lengthy outputs---occasionally even longer than the input text. While in some instances this detail may be preferred, we steer the model toward conciseness given our task of summarization.

Prompt phrasing and model temperature can have a considerable effect on LLM output, as demonstrated in the literature~\cite{strobelt2022interactive, wang2023prompt} and in Figure~\ref{fig:prompt-anatomy-engr}. For example, we achieve better performance by nudging the model towards expertise in medicine compared to expertise in wizardry or no specific expertise at all. This illustrates the value of relevant context in achieving better outcomes for the target task. We also explore the temperature hyperparameter, which adjusts the LLM’s conditional probability distributions during sampling, hence affecting how often the model will output less likely tokens. Higher temperatures lead to more randomness and ``creativity,'' while lower temperatures produce more deterministic outputs. Figure~\ref{fig:prompt-anatomy-engr} demonstrates that the lowest value, 0.1, performed best. We thus set temperature to this value for all models. Intuitively, a lower value seems appropriate given our goal of factually summarizing text with a high aversion to factually incorrect text.

~\subsubsection{Experimental Setup}\label{sec:experiment_details}

For each dataset, we construct test sets by randomly drawing the same $s$ samples, where $s=250$ for all datasets except dialogue ($s=100$), which includes only 126 samples in total. After selecting these $s$ samples, we choose another $s$ as a validation set for datasets which incorporated fine-tuning. We then use the remaining samples as a training set for ICL examples or QLoRA fine-tuning.

We leverage PyTorch for our all our experiments, including the parameter-efficient fine-tuning~\cite{peft} and the generative pre-trained transformers quantization~\cite{frantar2022gptq} libraries for implementing QLoRA. We fine-tune models with QLoRA for five epochs using the Adam optimizer with weight decay fix~\cite{loshchilov2017decoupled}. Our initial learning rate of $1 e^{-3}$ decays linearly to $1 e^{-4}$ after a 100-step warm-up; we determine this configuration after experimenting with different learning rates and schedulers. To achieve an effective batch size of 24 on each experiment, we adjust both individual batch size and number of gradient accumulation steps to fit on a single consumer GPU, a NVIDIA Quadro RTX 8000. 
All open-source models are available on HuggingFace~\cite{wolf2020transformers}.

~\subsubsection{Quantitative metrics}

We use well-known summarization metrics to assess the quality of generated summaries. BLEU~\cite{papineni2002bleu}, the simplest metric, calculates the degree of overlap between the reference and generated texts by considering 1- to 4-gram sequences. ROUGE-L~\cite{lin2004rouge} evaluates similarity based on the longest common subsequence; it considers both precision and recall, hence being more comprehensive than BLEU. In addition to these syntactic metrics, we employ BERTScore, which leverages contextual BERT embeddings to evaluate the semantic similarity of the generated and reference texts~\cite{zhang2019bertscore}. Lastly, we include MEDCON~\cite{yim2023aci} to gauge the consistency of medical concepts. This employs QuickUMLS~\cite{soldaini2016quickumls}, a tool that extracts biomedical concepts via string matching algorithms~\cite{okazaki2010simple}. We restrict MEDCON to specific UMLS semantic groups (Anatomy, Chemicals \& Drugs, Device, Disorders, Genes \& Molecular Sequences, Phenomena and Physiology) relevant for our work. All four metrics range from $[0, 100]$ with higher scores indicating higher similarity between the generated and reference summaries.

~\subsection{Clinical reader study}
After identifying the best model and method via NLP quantitative metrics, we perform a clinical reader study across three summarization tasks: radiology reports, patient questions, and progress notes. The dialogue task is excluded due to the unwieldiness of a reader parsing many lengthy transcribed conversations and paragraphs; see Figure~\ref{fig:clin-summ-example-d2n} for an example and Table~\ref{tab:datasets} for the token count.

Our readers include two sets of physicians: (1) five board-certified radiologists to evaluate summaries of radiology reports (2) five board-certified hospitalists (internal medicine physicians) to evaluate summaries of patient questions and progress notes. For each task, each physician views the same 100 randomly selected inputs and their A/B comparisons (\hhll{medical expert} vs.~\hhll{the best model} summaries), which are presented in a blinded and randomized order. An ideal summary would contain all clinically significant information (\textit{completeness}) without any errors (\textit{correctness}) or superfluous information (\textit{conciseness}). Hence we pose the following three questions for readers to evaluate using a five-point Likert scale.

\begin{itemize}
    \item \textbf{Completeness}: ``Which summary more completely captures important information?'' This compares the summaries' recall, i.e.~the amount of clinically significant detail retained from the input text.
    \item \textbf{Correctness}: ``Which summary includes less false information?'' This compares the summaries' precision, i.e.~instances of \hhll{fabricated information}. 
    \item \textbf{Conciseness}: ``Which summary contains less non-important information?'' This compares which summary is more condensed, as the value of a summary decreases with superfluous information.
\end{itemize}

Figure~\ref{fig:master-reader-study}e demonstrates the user interface for this study, which we create and deploy via Qualtrics. To obfuscate any formatting differences between the model and \hhll{medical expert} summaries, we apply simple post-processing to standardize capitalization, punctuation, newline characters, etc.

Given this non-parametric, categorical data, we assess the statistical significance of responses using a Wilcoxon signed-rank test with Type 1 error rate = 0.05 and adjust for multiple comparisons using the Bonferroni correction.
We estimate intra-reader correlation based on a mean-rating, fixed agreement, two-may mixed effects model~\cite{koo2016guideline} using the Pingouin package~\cite{vallat2018pingouin}. Additionally, readers \hhll{are provided comment space} to make observations for qualitative analysis.

~\subsection{Safety analysis}

\hhll{We conduct a safety analysis connecting summarization errors to medical harm, inspired by the Agency for Healthcare Research and Quality (AHRQ)’s harm scale}~\cite{walsh2017measuring}. This includes radiology reports ($n_r=27$) and progress notes ($n_n=44$) samples which contain disparities in completeness and/or correctness between the best model and medical expert summaries. Here, disparities occur if at least one physician significantly prefers or at least two physicians slightly prefer one summary to the other. These summary pairs are randomized and blinded. For each sample, we ask the following multiple-choice questions: ``Summary A is more complete and/or correct than Summary B. Now, suppose Summary B (worse) is used in the standard clinical workflow. Compared to using Summary A (better), what would be the...'' (1) ``... extent of possible harm?'' options: \{none, mild or moderate harm, severe harm or death\} (2) ``... likelihood of possible harm?'' options: \{low, medium, high\}.

Safety analysis of fabricated information is discussed in Section~\ref{sec:results_correctness}.

\subsection{Connecting quantitative and clinical evaluations}\label{sec:corr}

We now provide intuition connecting NLP metrics and clinical reader scores. Note that in our work, these tools measure different quantities; NLP metrics measure the similarity between two summaries, while reader scores measure which summary is better. Consider an example where two summaries are exactly the same: NLP metrics would yield the highest possible score (100), while clinical readers would provide a score of 0 to denote equivalence. As the magnitude of a reader score increases, the two summaries are increasingly dissimilar, hence yielding a lower quantitative metric score. Given this intuition, we compute the Spearman correlation coefficient between NLP metric scores and the magnitude of the reader scores. Since these features are inversely correlated, for clarity we display the negative correlation coefficient values.
\begin{figure*}[t!]
    \centering
    \includegraphics[width = 1\textwidth]{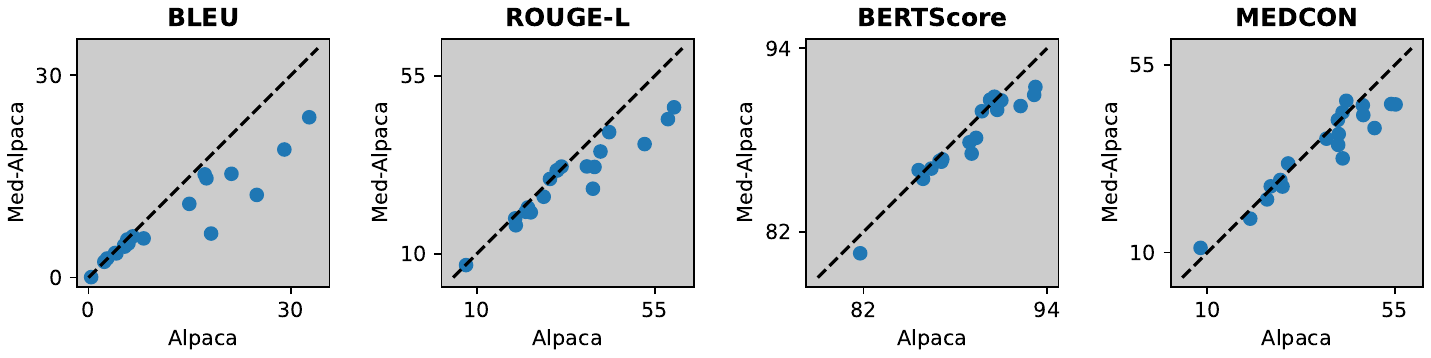}
    \caption{\textbf{Alpaca vs.~Med-Alpaca.} Given that most data points are below the dashed lines denoting equivalence, we conclude that Med-Alpaca's fine-tuning with medical Q\&A data results in worse performance for our clinical summarization tasks. See Section~\ref{sec:results_quant_eval} for further discussion. Note that each data point corresponds to the average score of $s=250$ samples for a given experimental configuration, i.e.~\{dataset $\times$ $m$ in-context examples\}.
    }
    \label{fig:yx_alpaca}
\end{figure*}
\section{Results and Discussion}

\subsection{Quantitative evaluation}\label{sec:results_quant_eval}

\subsubsection{Impact of domain-specific fine-tuning}

When considering which open-source models to evaluate, we first assess the benefit of fine-tuning open-source models on medical text. For example, Med-Alpaca~\cite{han2023medalpaca} is a version of Alpaca~\cite{alpaca} which was further instruction-tuned with medical Q\&A text, consequently improving performance for the task of medical question-answering. Figure~\ref{fig:yx_alpaca} compares these models for our task of summarization, showing that most data points are below the dashed lines denoting equivalence. Hence despite Med-Alpaca's adaptation for the medical domain, it performs worse than Alpaca for our tasks of clinical text summarization---\hhll{highlighting a distinction between domain adaptation and task adaptation.}
With this in mind, and considering that Alpaca is commonly known to perform worse than our other open-source autoregressive models Vicuna and Llama-2~\cite{zheng2023judging, vicuna2023}, for simplicity we exclude Alpaca and Med-Alpaca from further analysis.

\begin{figure*}[b!]
    \centering
    \includegraphics[width = 0.85\textwidth]{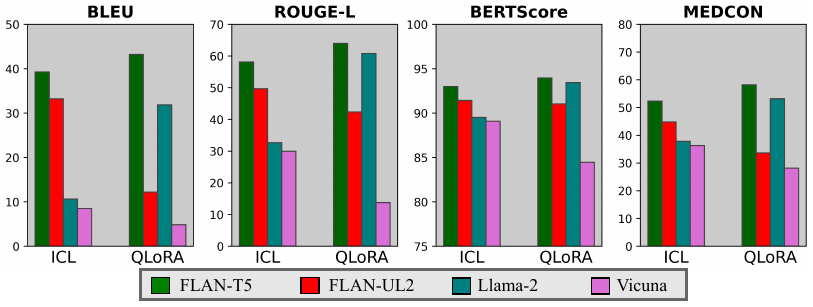}
    \caption{
    \textbf{One in-context example (ICL) vs. QLoRA} across open-source models on Open-i radiology reports. FLAN-T5 achieves best performance on both methods for this dataset. While QLoRA typically outperforms ICL with the better models (FLAN-T5, Llama-2), this relationship reverses given sufficient in-context examples (Figure~\ref{fig:grid-of-graphs}).
    Figure~\ref{fig:icl-v-lora-chq} contains similar results with patient health questions.
    }
    \label{fig:icl-v-lora-opi}
\end{figure*}
\begin{figure*}[t!]
    \centering
    \includegraphics[width = 1\textwidth]{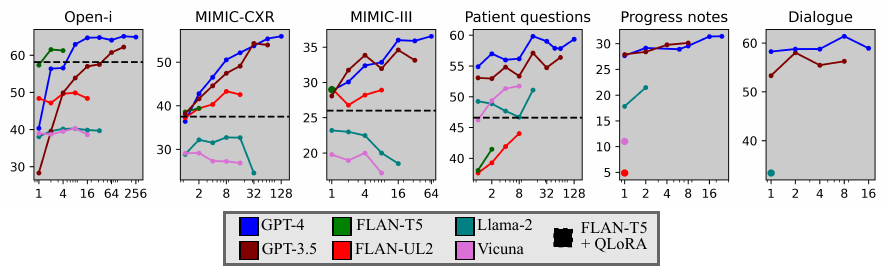}
    \caption{
    \textbf{MEDCON scores vs. number of in-context examples} across models and datasets. We also include the best model fine-tuned with QLoRA (FLAN-T5) as a horizontal dashed line for valid datasets. Zero-shot prompting (0 examples) often yields considerably inferior results, underscoring the need for adaptation methods. Note the allowable number of in-context examples varies significantly by model and dataset. See Figure~\ref{fig:grid-of-graphs} for results across all four metrics.}
    \label{fig:graphs-medcon}
\end{figure*}

\subsubsection{Comparison of adaptation strategies}

Next, we compare ICL (in-context learning) vs.~QLoRA (quantized low-rank adaptation) across the remaining open-source models using the Open-i radiology report dataset in Figure~\ref{fig:icl-v-lora-opi} and the patient health questions in Figure~\ref{fig:icl-v-lora-chq}. We choose these datasets because their shorter context lengths allow for training with lower computational cost. FLAN-T5 emerged as the best-performing model with QLoRA. QLoRA typically outperformed ICL (one example) with the better models FLAN-T5 and Llama-2\hhll{; given a sufficient number of in-context examples, however, most models surpass even the best QLoRA fine-tuned model, FLAN-T5 (}Figure~\ref{fig:grid-of-graphs}). FLAN-T5 (2.7B) eclipsed its fellow seq2seq model FLAN-UL2 (20B), despite being an older model with almost 8$\times$ fewer parameters.

When considering trade-offs between adaptation strategies, availability of these models (open-source vs. proprietary) raises an interesting consideration for healthcare, where data and model governance are important---especially if summarization tools are cleared for clinical use by the Food and Drug Administration. This could motivate the use of fine-tuning methods on open-source models. Governance aside, ICL provides many benefits: (1) model weights are fixed, hence enabling queries of pre-existing LLMs (2) adaptation is feasible with even a few examples, while fine-tuning methods such as QLoRA typically require hundreds or thousands of examples.

\subsubsection{Effect of context length for in-context learning}

Figure~\ref{fig:graphs-medcon} displays MEDCON~\cite{yim2023aci} scores for all models against number of in-context examples, up to the maximum number of allowable examples for each model and dataset. This graph also includes the best performing model (FLAN-T5) with QLoRA as a reference, depicted by a horizontal dashed line. Compared to zero-shot prompting ($m=0$ examples), adapting with even $m=1$ example considerably improves performance in almost all cases, underscoring the importance of adaptation methods.
While ICL and QLoRA are competitive for open-source models, proprietary models GPT-3.5 and GPT-4 far outperform other models and methods given sufficient in-context examples.
For a similar graph across all metrics, see Figure~\ref{fig:grid-of-graphs}.

\subsubsection{Head-to-head model comparison}

\begin{figure*}[t]
    \centering
    \includegraphics[width = 1\textwidth]{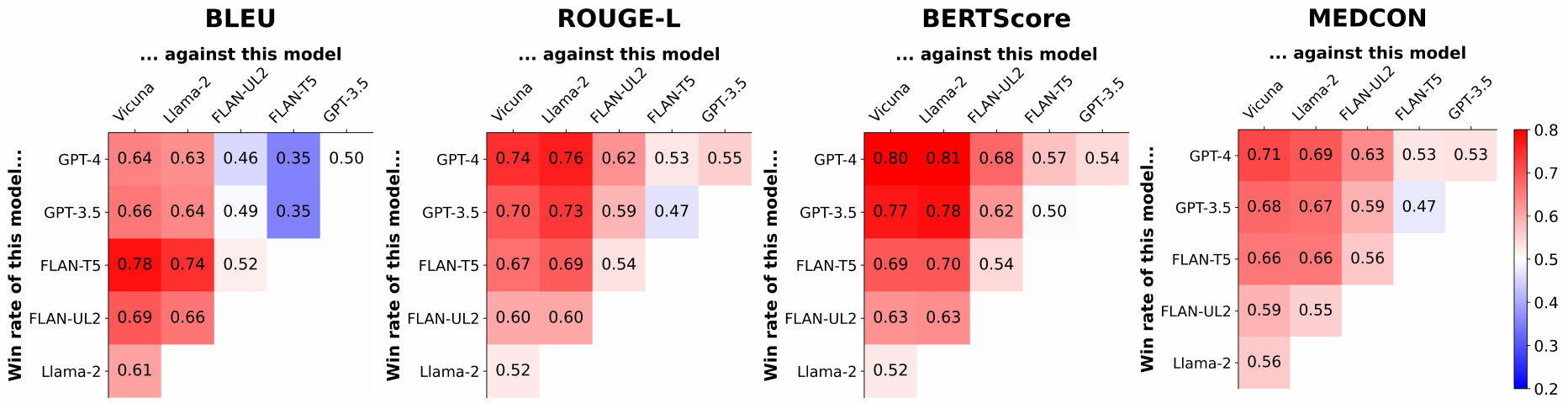}
    \caption{\textbf{Model win rate}: a head-to-head winning percentage of each model combination, where red/blue intensities highlight the degree to which models on the vertical axis outperform models on the horizontal axis. GPT-4 generally achieves the best performance. While FLAN-T5 is more competitive for syntactic metrics such as BLEU, we note this model is constrained to shorter context lengths (Table~\ref{tab:models}). When aggregated across datasets, seq2seq models (FLAN-T5, FLAN-UL2) outperform open-source autoregressive models (Llama-2, Vicuna) on all metrics.}
    \label{fig:win-matrix-main}
\end{figure*}

Figure~\ref{fig:win-matrix-main} compares models using win rates, i.e.~the head-to-head winning percentage of each model combination across the same set of samples. In other words, for what percentage of samples do model A's summaries have a higher score than model B's summaries? This presents trade-offs of different model types. Seq2seq models (FLAN-T5, FLAN-UL2) perform well on syntactical metrics such as BLEU~\cite{papineni2002bleu} but worse on others, suggesting that these models excel more at matching word choice than matching semantic or conceptual meaning. Note seq2seq models are often constrained to much shorter context length than autoregressive models (Table~\ref{tab:models}), because seq2seq models require the memory-intensive step of encoding the input sequence into a fixed-size context vector. Among open-source models, seq2seq models perform better than autoregressive (Llama-2, Vicuna) models on radiology reports but worse on patient questions and progress notes (Figure~\ref{fig:grid-of-graphs}). Given that these latter datasets have higher lexical variance (Table~\ref{tab:datasets}) and more heterogeneous formatting compared to radiology reports, we hypothesize that autoregressive models may perform better with increasing data heterogeneity and complexity.

\hhll{\textbf{Best model/method.}} We deemed the best model and method to be GPT-4 (context length 32,768) with a maximum allowable number of in-context examples, \hhll{hereon identified as the best-performing model.}

\subsection{Clinical reader study} 

Given our clinical reader study design (Figure~\ref{fig:master-reader-study}a), pooled results across ten physicians (Figure~\ref{fig:master-reader-study}b) demonstrate that summaries from~\hhll{the best adapted model (GPT-4 using ICL)} are more complete and contain fewer errors compared to medical expert summaries---which were created either by medical doctors during clinical care or by a committee of medical doctors and experts.

The distributions of reader responses in Figure~\ref{fig:master-reader-study}c show that medical expert summaries are preferred in only a minority of cases (19\%), while in a majority, the best model is either non-inferior (45\%) or preferred (36\%). Table A1 contains scores separated by individual readers and affirms the reliability of scores across readers by displaying positive intra-reader correlation values. Based on physician feedback, we undertake a qualitative analysis to illustrate strengths and weaknesses of summaries by the model and medical experts; see Figures~\ref{fig:clin-summ-example-iii},~\ref{fig:clin-summ-example-chq}, and~\ref{fig:clin-summ-example-pls}. Now, we discuss results with respect to each individual attribute.

\subsubsection{Completeness}\label{sec:results_completeness}
\begin{figure}[h!] % must use h! to avoid figure being placed at end of document
    \centering
    \includegraphics[width = 1\textwidth]{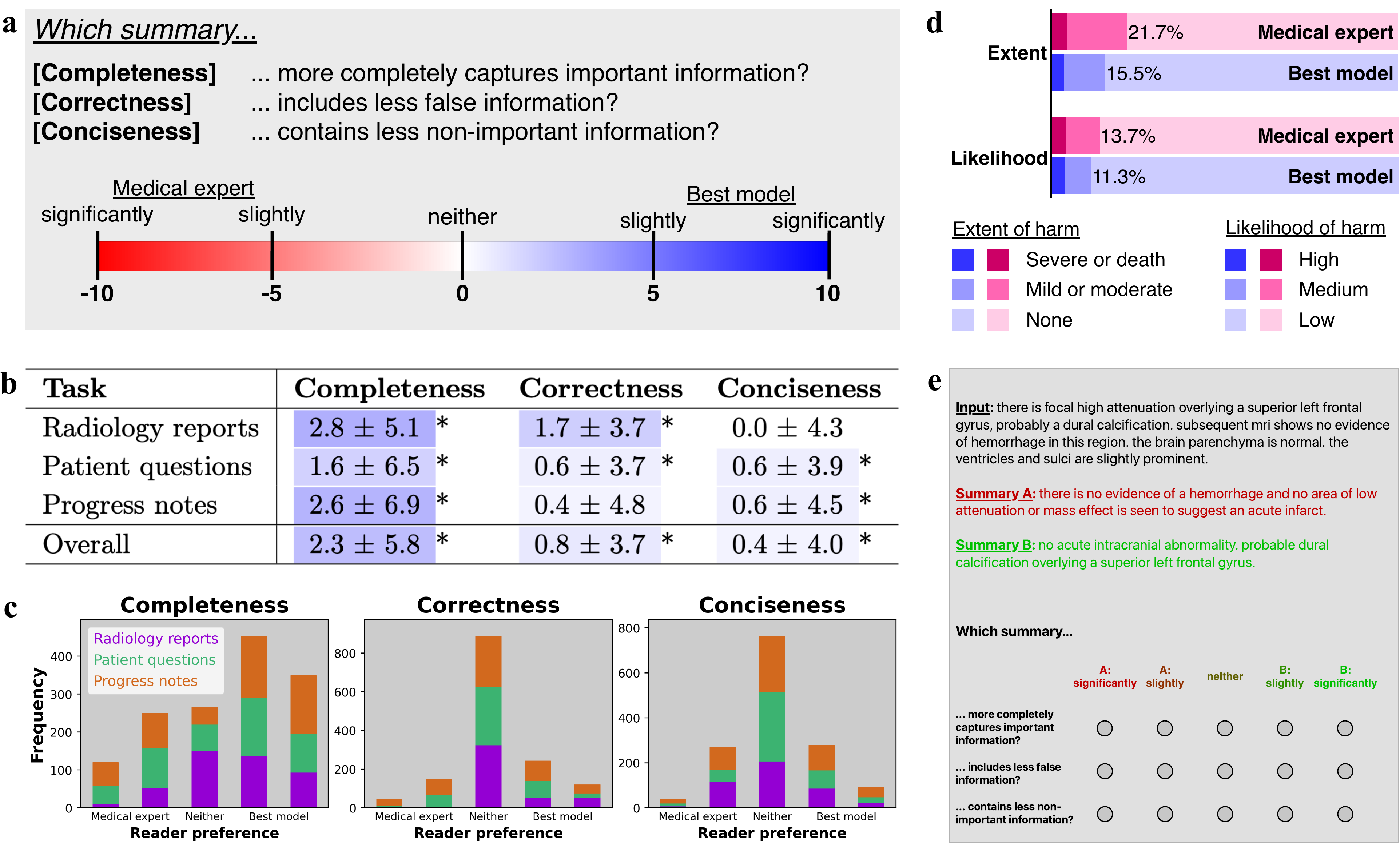}
    \caption{
    \textbf{Clinical reader study.}
    \textbf{(a)}
    Study design comparing the summaries from the best model versus that of medical experts on three attributes: completeness, correctness, and conciseness.
    \textbf{(b)} Results. Model summaries are rated higher on all attributes.
    Highlight colors correspond to a value's location on the color spectrum. Asterisks (*) denote statistical significance by Wilcoxon signed-rank test, $p < 0.001$.
    \textbf{(c)}
    Distribution of reader scores.
    Horizontal axes denote reader preference as measured by a five-point Likert scale.
    Vertical axes denote frequency count, with 1,500 total reports for each plot.
    \textbf{(d)}
    \hhll{Extent and likelihood of potential medical harm caused by choosing summaries from the medical expert (pink) or best model (purple) over the other. Model summaries are preferred in both categories.}
    \textbf{(e)}
    Reader study user interface. 
    }
    \label{fig:master-reader-study}
\end{figure}

The best model summaries are more complete on average than medical expert summaries, achieving statistical significance across all three summarization tasks with $p < 0.001$ (Figure~\ref{fig:master-reader-study}b). \hhll{Lengths of summaries were comparable between the model and medical experts for all three datasets: $47\pm24$ vs. $44\pm22$ tokens for radiology reports, $15\pm5$ vs. $14\pm4$ tokens for patient questions, and $29\pm7$ vs. $27\pm13$ tokens for progress notes (all $p > 0.12$). Hence the model's advantage in completeness is not simply a result of generating longer summaries.}
We provide intuition for completeness by investigating a specific example in progress notes summarization. In Figure~\ref{fig:clin-summ-example-pls}, \hhll{the model} correctly identifies conditions that were missed by the medical expert, such as hypotension and anemia. \hhll{Although the model} was more complete in generating its progress notes summary, \hhll{it} also missed historical context (a history of HTN, or hypertension).

\subsubsection{Correctness}\label{sec:results_correctness}
\begin{figure*}%[b!]
    \centering
    \includegraphics[width = 1.\textwidth]{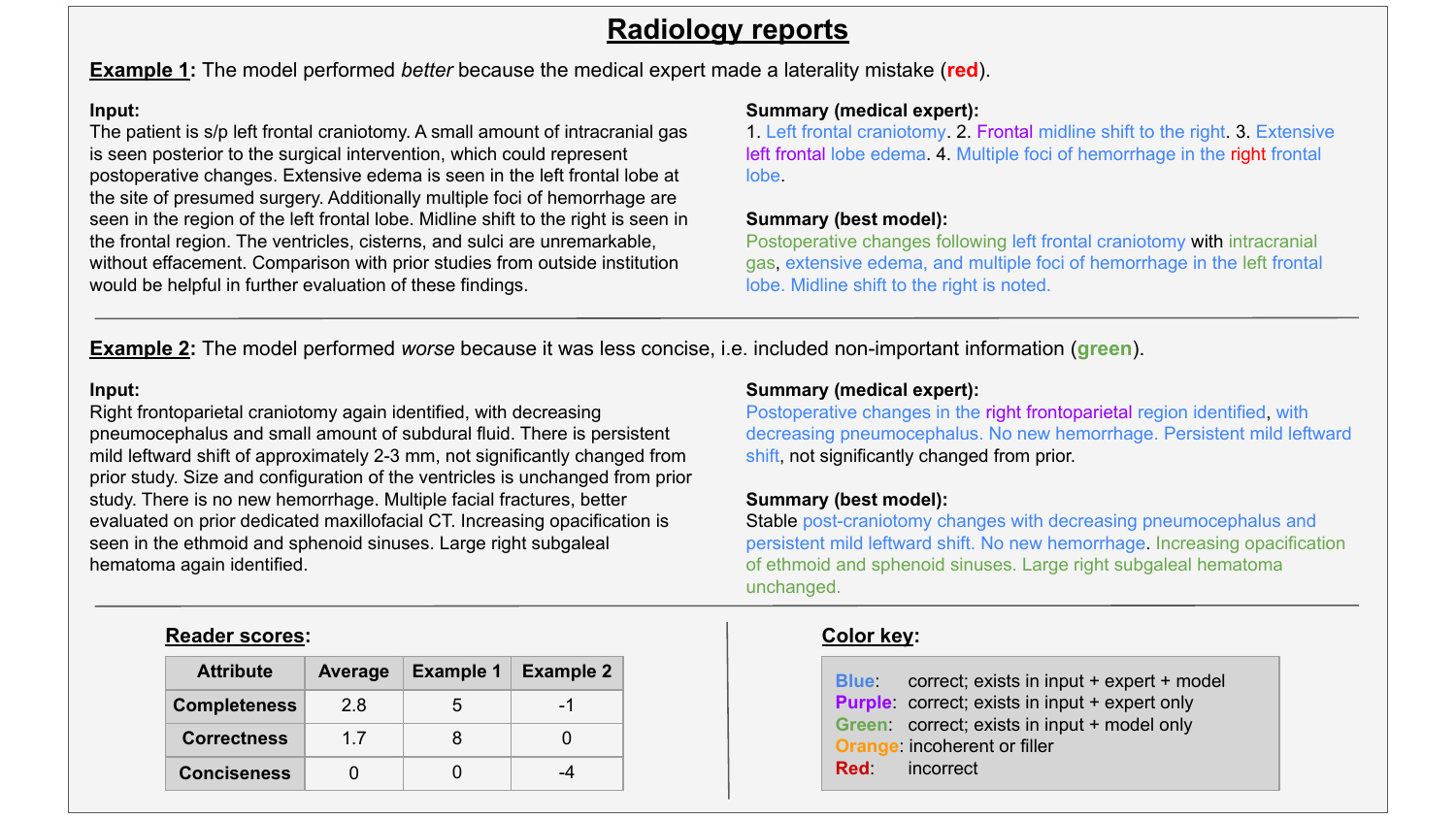}
    \caption{\textbf{Annotation: radiology reports.} The table (lower left) contains reader scores for these two examples and the task average across all samples. \underline{\smash{Top}}: the model performs better due to a laterality mistake by the medical expert. \underline{\smash{Bottom}}: the model exhibits a lack of conciseness. 
    }
    \label{fig:clin-summ-example-iii}
\end{figure*}

With regards to correctness, \hhll{the best model} generated significantly fewer errors ($p < 0.001$) compared to \hhll{medical} expert summaries overall and on two of three summarization tasks (Figure~\ref{fig:master-reader-study}b).
As an example of the \hhll{model}'s superior correctness performance on radiology reports, we observe that it avoided common~\hhll{medical expert} errors related to lateral distinctions (right vs. left, Figure \ref{fig:clin-summ-example-iii}).
For progress notes, Figure~\ref{fig:clin-summ-example-pls} reveals an intriguing case: during the blinded study, the physician reader erroneously assumed that a hallucination---the incorrect inclusion of a urinary tract infection---was made by the model. In this case, the medical expert was responsible for the hallucination. This instance underscores the point that even medical experts, not just LLMs, can hallucinate.
Despite this promising performance, the model was not perfect across all tasks. We see a clear example in Figure~\ref{fig:clin-summ-example-pls} in which \hhll{the model} mistakenly generated several conditions in the problem list that were incorrect, such as eosinophilia.

Both the model and medical experts faced challenges interpreting ambiguity, such as user queries in patient health questions. Consider Figure~\ref{fig:clin-summ-example-chq}'s first example, in which the input question mentioned ``diabetes and neuropathy.''  The model mirrored this phrasing verbatim, while the medical expert interpreted it as ``diabetic neuropathy.'' In Figure~\ref{fig:clin-summ-example-chq}'s second example, the model simply reformulated the input question about tests and their locations, while the medical expert inferred a broader query about tests and treatments. In both cases, the model’s summaries leaned toward literalness, a trait that readers sometimes favored and sometimes did not. In future work, a systematic exploration of model temperature could further illuminate this trade-off.

Further, the critical need for accuracy in a clinical setting motivates a more nuanced understanding of correctness. As such, we define three types of fabricated information: (1) misinterpretations of ambiguity, (2) factual inaccuracies: modifying existing facts to be incorrect, and (3) hallucinations: inventing new information that cannot be inferred from the input text. We found that the model committed these errors on 6\%, 2\%, and 5\% of samples, respectively, compared to 9\%, 4\%, and 12\% by medical experts. Given the model’s lower error rate in each category, this suggests that incorporating LLMs could actually reduce \hhll{fabricated information} in clinical practice.

Beyond the scope of our work, there's further potential to reduce fabricated information through incorporating checks by a human, checks by another LLM, or using a model ensemble to create a ``committee of experts''~\cite{jozefowicz2016exploring, chang2023survey}.

\subsubsection{Conciseness}
With regards to conciseness, \hhll{the best model} performed significantly better ($p < 0.001$) overall and on two \hhll{tasks (Figure~\ref{fig:master-reader-study}b). We note the model's summaries are more concise while concurrently being more complete.} Radiology reports were the only task in which physicians did not prefer the best model's summaries to medical experts. See Figure~\ref{fig:clin-summ-example-iii} for an example. We suggest that conciseness could be improved with better prompt engineering, or modifying the prompt to improve performance. Of the task-specific instructions in Table~\ref{tab:datasets}, the other two tasks (patient questions, progress notes) explicitly specify summary length, e.g.~``15 words or less.'' These phrases are included so that model summaries are generated with similar lengths to the human summaries, enabling a clean comparison. Length specification in the radiology reports prompt instruction was more vague, i.e.~``...with minimal text,'' perhaps imposing a softer constraint on the model. We leave further study of prompt instructions to future work.  

\subsection{Safety Analysis}\label{sec:results_fab_info}
\begin{figure*}[t]
    \centering
    \includegraphics[width = 0.55\textwidth]{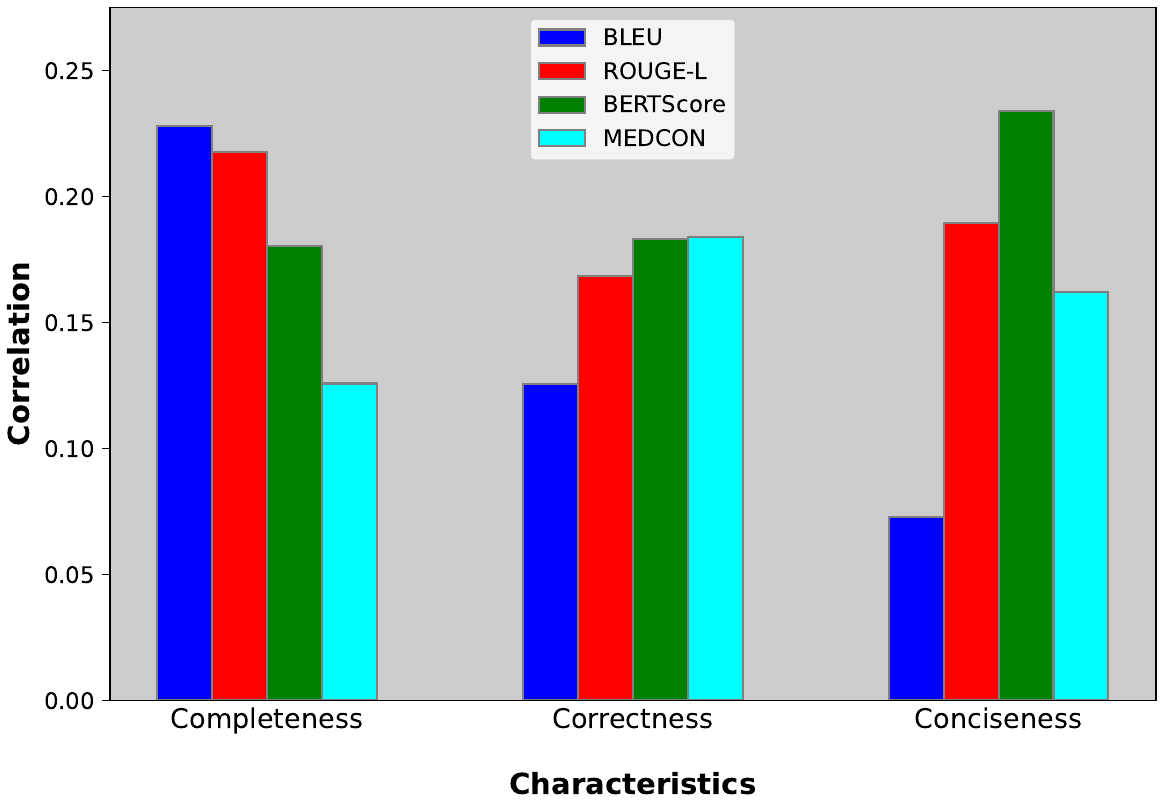}
    \caption{
    \textbf{Correlation between NLP metrics and reader scores.} The semantic metric (BERTScore) and conceptual metric (MEDCON) correlate most highly with correctness. Meanwhile, syntactic metrics BLEU and ROUGE-L correlate most with completeness. See Section~\ref{sec:corr} for further discussion.
    }
    \label{fig:corr-plot}
\end{figure*}

The results of this harm study (Figure~\ref{fig:master-reader-study}d) indicate that the medical expert summaries would have both a higher likelihood (14\%) and higher extent (22\%) of possible harm compared to the summaries from the best model (12\% and 16\%, respectively). These percentages are computed with respect to all samples, such that the subset of samples with similar A/B summaries (in completeness and correctness) are assumed to contribute no harm. For the safety analysis of fabricated information, please see Section~\ref{sec:results_correctness}. Ultimately we argue that, beyond clinical reader studies, conducting downstream analyses is crucial to affirm the safety of LLM-generated summaries in clinical environments.

\subsection{Connecting quantitative and clinical evaluations}\label{sec:corr}

Figure~\ref{fig:corr-plot} captures the correlation between NLP metrics and physicians' preference. Compared to other metrics, BLEU correlates most with completeness and least with conciseness. Given that BLEU measures sequence overlap, this result seems reasonable, as more text provides more ``surface area'' for overlap; more text also reduces the brevity penalty that BLEU applies on generated sequences which are shorter than the reference~\cite{papineni2002bleu}. The metrics BERTScore (measuring semantics) and MEDCON (measuring medical concepts) correlate most strongly with reader preference for correctness. Overall, however, the low magnitude of correlation values (approximately 0.2) underscores the need to go beyond NLP metrics \hhll{with a reader study} when assessing clinical readiness.

Aside from the low correlation values in Figure~\ref{fig:corr-plot}, our reader study results (Figure~\ref{fig:master-reader-study}b) highlight another limitation of NLP metrics, especially as model-generated summaries become increasingly viable. These metrics rely on a reference---in our case, medical expert summaries---which we have demonstrated may contain errors. Hence we suggest that human evaluation is essential when assessing the clinical feasibility of new methods. If human evaluation is not feasible, Figure~\ref{fig:corr-plot} suggests that syntactic metrics are better at measuring completeness, while semantic and conceptual metrics are better at measuring correctness.

\subsection{Limitations}

This study has several limitations which motivate future research.

Model temperature and prompt phrasing can be important for LLM performance (Figure~\ref{fig:prompt-anatomy-engr}),~\cite{strobelt2022interactive, wang2023prompt}. However, we only search over three possible temperature values. Further, we do not thoroughly engineer our prompt instructions (Table~\ref{tab:datasets}); each was chosen after trying only 1-2 options over a small dataset. While this highlights the potential for improvement, we're also encouraged that achieving convincing results does not require a thorough temperature search or prompt engineering.

In our quantitative analysis, we select state-of-the-art and highly regarded LLMs with a diverse range of attributes. This includes the 7B-parameter tier of open-source autoregressive models, despite some models such as Llama-2 having larger versions. We consider the benefit of larger models in Figure~\ref{fig:yx_llama2}, finding this improvement marginal for Llama-2 (13B) compared to Llama-2 (7B). While there may exist open-source models which perform slightly better than our selections, we do not believe this would meaningfully alter our analysis---especially considering the clinical reader study employs GPT-4, which is an established state-of-the-art~\cite{zheng2023judging}.

Our study does not encompass all clinical document types, and extrapolating our results is tentative. For instance, our progress notes task employs ICU notes from a single medical center. These notes may be structured differently from non-ICU notes or from ICU notes of a different center. Additionally, more challenging tasks may require summarizing longer documents or multiple documents of different types. Addressing these cases demands two key advancements: (1) extending model context length, potentially through multi-query aggregation or other methods~\cite{poli2023hyena, ding2023longnet} (2) introducing open-source datasets that include broader tasks and lengthier documents.
We thus advocate for expanding evaluation to other summarization tasks.

We do not consider the inherently context-specific nature of summarization. For example, a gastroenterologist, radiologist, and oncologist may have different preferences for summaries of a cancer patient with liver metastasis. Or perhaps an abdominal radiologist will want a different summary than a neuroradiologist. Further, individual clinicians may prefer different styles or amounts of information. While we do not explore such a granular level of adaptation, this may not require much further development: since the best model and method uses a handful of examples via ICL, one could plausibly adapt using examples curated for a particular specialty or clinician. Another limitation is that radiology report summaries from medical experts occasionally recommend further studies or refer to prior studies, e.g. ``... not significantly changed from prior'' in Figure~\ref{fig:clin-summ-example-iii}. These instances are out of scope for our tasks, which do not include context from prior studies; hence in the clinical reader study, physicians were told to disregard these phrases. Future work can explore providing the LLM with additional context and longitudinal information.

An additional consideration for ours and other LLM studies, especially with proprietary models, is that it is not possible to verify whether a particular open-source dataset was included in model training. While three of our datasets (MIMIC-CXR, MIMIC-III, ProbSum) require PhysioNet~\cite{PhysioNet} access to ensure safe data usage by third parties, this is no guarantee against data leakage. This complication highlights the need for validating results on internal data when possible.

We note the potential for LLMs to be biased~\cite{omiye2023large, zack2024assessing}. While our datasets do not contain demographic information, we advocate for future work to consider whether summary qualities have any dependence upon group membership.
\section{Conclusion}

In this research, we evaluate methods for adapting LLMs to summarize clinical text, analyzing eight models across a diverse set of summarization tasks. Our quantitative results underscore the advantages of adapting models to specific tasks and domains. The ensuing clinical reader study demonstrates that LLM summaries are often preferred over medical expert summaries due to higher scores for completeness, correctness, and conciseness. The subsequent safety analysis explores qualitative examples, potential medical harm, and fabricated information to demonstrate the limitations of both LLMs and medical experts. Evidence from this study suggests that incorporating LLM-generated candidate summaries into the clinical workflow could reduce documentation load, potentially leading to decreased clinician strain and improved patient care. Testing this hypothesis motivates future prospective studies in clinical environments.

\section{Acknowledgements}
Microsoft provided Azure OpenAI credits for this project via both the Accelerate Foundation Models Academic Research (AFMAR) program and also a cloud services grant to Stanford Data Science. Further compute support was provided by One Medical, which Asad Aali used as part of his summer internship. Curtis Langlotz is supported by NIH grants R01 HL155410, R01 HL157235, by AHRQ grant R18HS026886, by the Gordon and Betty Moore Foundation, and by the National Institute of Biomedical Imaging and Bioengineering (NIBIB) under contract 75N92020C00021. Akshay Chaudhari receives support from NIH grants R01 HL167974, R01 AR077604, R01 EB002524, R01 AR079431, and P41 EB027060; from NIH contracts 75N92020C00008 and 75N92020C00021; and from GE Healthcare, Philips, and Amazon.

\section{Data and Code Availability}
While all datasets are publicly available, our GitHub repository \url{github.com/StanfordMIMI/clin-summ} includes preprocessed versions for those which do not require PhysioNet access: Open-i~\cite{demner2016preparing} (radiology reports), MeQSum~\cite{MeQSum} (patient questions), and ACI-Bench~\cite{yim2023aci} (dialogue). Researchers can also access the original datasets via the provided references. Any further distribution of datasets is subject to the terms of use and data sharing agreements stipulated by the original creators. Our repository also contains experiment code and links to open-source models hosted by HuggingFace~\cite{wolf2020transformers}.

\section{Author contributions}
DVV collected data, developed code, ran experiments, designed studies, analyzed results, created figures, and wrote the manuscript. All authors reviewed the manuscript, providing meaningful revisions and feedback. CVU, LB, JBD provided technical advice in addition to conducting qualitative analysis (CVU), building infrastructure for the Azure API (LB), and implementing the MEDCON metric (JB). AA assisted in model fine-tuning. CB, AP, MP, EPR, AS participated in the reader study as radiologists. NR, PH, WC, NA, JH participated in the reader study as hospitalists. CPL, JP, ASC provided student funding. SG advised on study design for which JH and JP provided additional feedback. JP, ASC guided the project, with ASC serving as principal investigator and advising on technical details and overall direction. No funders or third parties were involved in study design, analysis, or writing.

\clearpage
\setlength\bibitemsep{3pt}
\printbibliography
\clearpage

\clearpage
% \clearpage
\appendix
\section{Appendix}
\label{sec:appendix}

\setcounter{figure}{0}  
\renewcommand{\thefigure}{A\arabic{figure}}
\setcounter{table}{0}  
\renewcommand{\thetable}{A\arabic{table}}

\vspace{-5mm}
\begin{figure*}[b!]
    \centering
    \includegraphics[width = 0.85\textwidth]{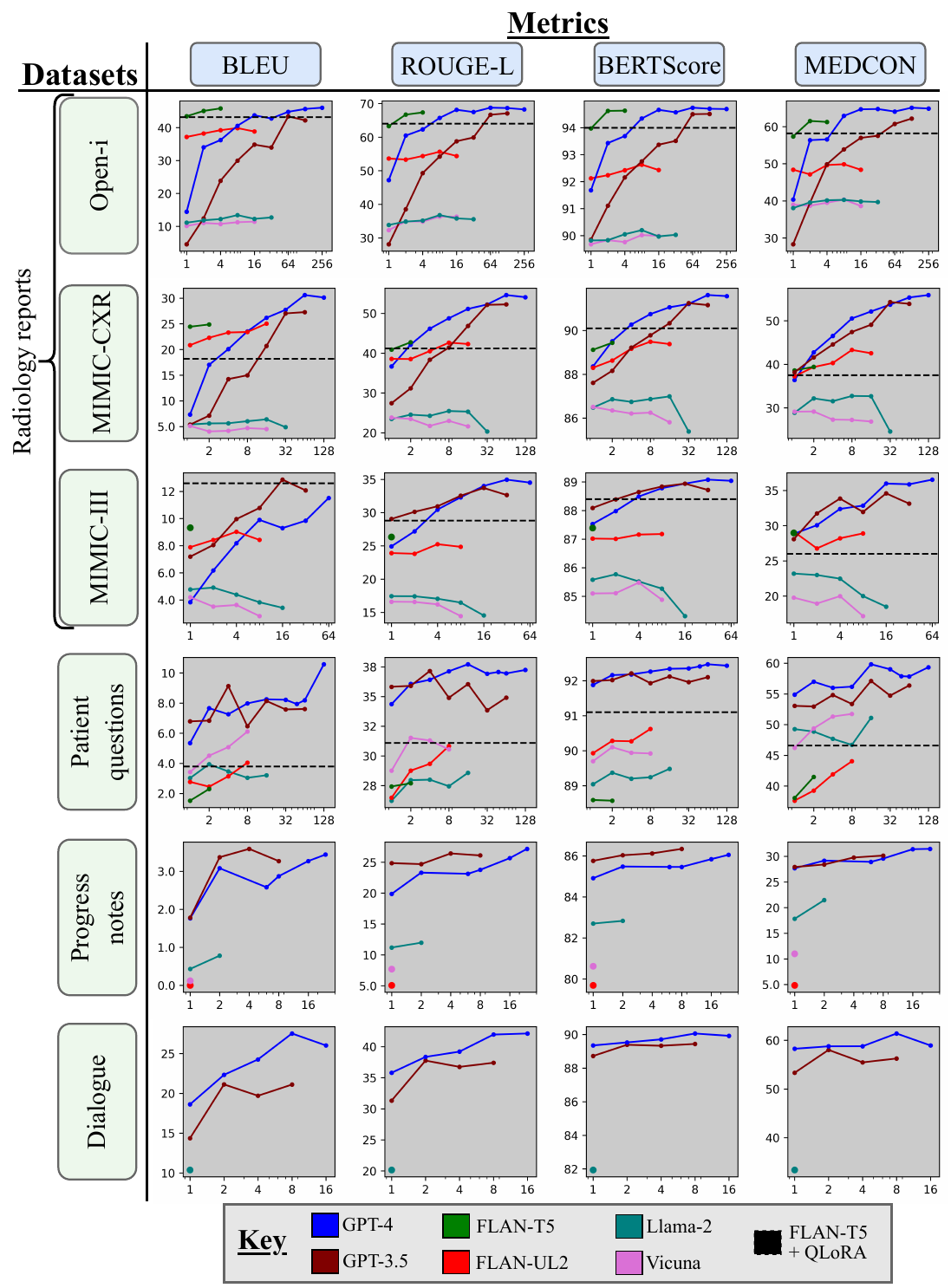}
    \caption{
    Metric scores vs. number of in-context examples across models and datasets. We also include the best model fine-tuned with QLoRA (FLAN-T5) as a horizontal dashed line. Note the allowable number of in-context examples varies significantly by model and dataset.}
    \label{fig:grid-of-graphs}
\end{figure*}
\begin{figure*}[t]
    \centering
    \includegraphics[width = 1\textwidth]{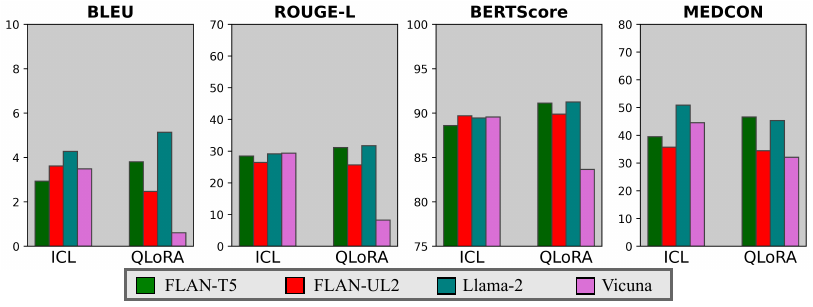}
    \caption{
    One in-context example (ICL) vs. QLoRA across open-source models on patient health questions. While QLoRA typically outperforms ICL with the better models (FLAN-T5, Llama-2), this relationship reverses given sufficient in-context examples (Figure~\ref{fig:grid-of-graphs}). Figure~\ref{fig:icl-v-lora-opi} contains similar results with the Open-i radiology report dataset.
    }
    \label{fig:icl-v-lora-chq}
\end{figure*}
\begin{figure*}[t]
    \centering
    \includegraphics[width = 1\textwidth]{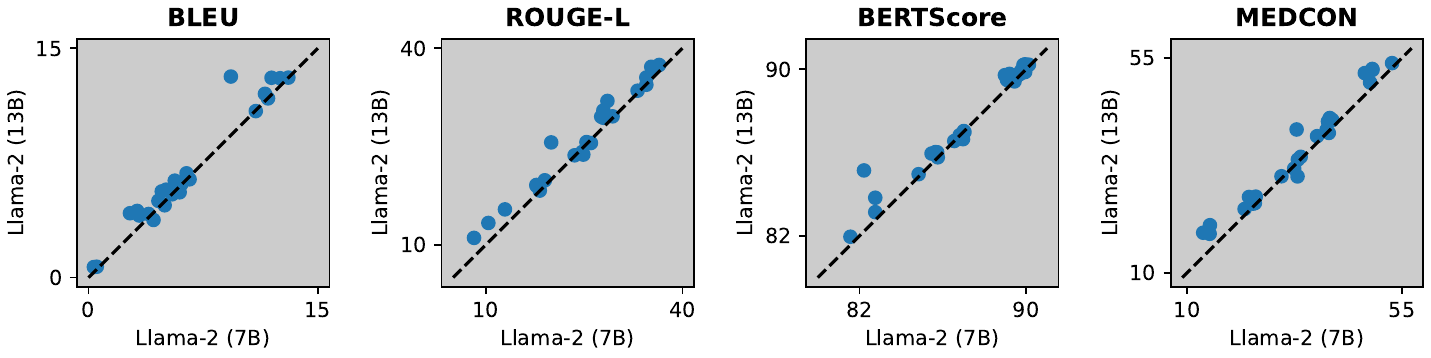}
    \caption{Comparing Llama-2 (7B) vs. Llama-2 (13B). As most data points are near or slightly above the dashed lines denoting equivalence, we conclude that the larger Llama-2 model (13B parameters) delivers marginal improvement for clinical summarization tasks compared to the 7B model. Note that each data point corresponds to the average score of $s=250$ samples for a given experimental configuration, i.e.~\{dataset $\times$ $m$ in-context examples\}.
    }
    \label{fig:yx_llama2}
\end{figure*}
\clearpage

\setlength{\tabcolsep}{6pt} % spacing b/w cols, default 6pt
\setlength{\extrarowheight}{1.5pt}

\begin{table*}[h]
  {\caption{Reader study results evaluating completeness, correctness, conciseness (columns) across individual readers \hhll{and pooled across readers}. Scores are on the range [-10, 10], where positive scores denote \hhll{the best model} is preferred to the \hhll{medical expert}. Intensity of highlight colors blue (\hhll{model} wins) or red (\hhll{expert} wins) correspond to the score. \hhll{Asterisks (*) on pooled rows denote statistical significance by a one-sided Wilcoxon signed-rank test, $p < 0.001$. Intra-class correlation (ICC) values across readers are on a range of $[-1, 1]$ where $-1$, $0$, and $+1$ correspond to negative, no, and positive correlations, respectively.} See Figure~\ref{fig:master-reader-study}\hhll{a} for \hhll{study overview}.}}
  \begin{center} 

{\begin{tabular}{c|c|lll}
  \hline
  \textbf{Task} & \textbf{Reader}
              & \textbf{Completeness} & \textbf{Correctness} & \textbf{Conciseness} \\
 \hline 
 & 1 & \colorbox{d1_r1_q1}{\makebox[\cboxlength][c]{3.5 $\pm$ 5.6}} & \colorbox{d1_r1_q2}{\makebox[\cboxlength][c]{1.7 $\pm$ 3.6}} & \colorbox{d1_r1_q3}{\makebox[\cboxlength][c]{1.2 $\pm$ 4.8}} \\ 
 & 2 & \colorbox{d1_r2_q1}{\makebox[\cboxlength][c]{3.6 $\pm$ 6.6}} & \colorbox{d1_r2_q2}{\makebox[\cboxlength][c]{2.5 $\pm$ 4.7}} & \colorbox{d1_r2_q3}{\makebox[\cboxlength][c]{-0.3 $\pm$ 5.4}} \\ 
 \multirow{2}{*}{\textbf{Radiology}} & 3 & \colorbox{d1_r3_q1}{\makebox[\cboxlength][c]{0.8 $\pm$ 2.9}} & \colorbox{d1_r3_q2}{\makebox[\cboxlength][c]{0.6 $\pm$ 3.2}} & \colorbox{d1_r3_q3}{\makebox[\cboxlength][c]{-1.7 $\pm$ 3.0}} \\ 
 \multirow{2}{*}{\textbf{reports}} & 4 & \colorbox{d1_r4_q1}{\makebox[\cboxlength][c]{4.7 $\pm$ 4.7}} & \colorbox{d1_r4_q2}{\makebox[\cboxlength][c]{2.9 $\pm$ 3.9}} & \colorbox{d1_r4_q3}{\makebox[\cboxlength][c]{1.2 $\pm$ 3.8}} \\ 
 & 5 & \colorbox{d1_r5_q1}{\makebox[\cboxlength][c]{1.4 $\pm$ 4.0}} & \colorbox{d1_r5_q2}{\makebox[\cboxlength][c]{0.6 $\pm$ 2.2}} & \colorbox{d1_r5_q3}{\makebox[\cboxlength][c]{-0.6 $\pm$ 3.4}} \\ 
 & Pooled & \colorbox{d1_ra_q1}{\makebox[\cboxlength][c]{2.8 $\pm$ 5.1}}* & \colorbox{d1_ra_q2}{\makebox[\cboxlength][c]{1.7 $\pm$ 3.7}}* & \colorbox{d1_ra_q3}{\makebox[\cboxlength][c]{0.0 $\pm$ 4.3}} \\
 & ICC & \makebox[\cboxlength][c]{0.45} & \makebox[\cboxlength][c]{0.58} & \makebox[\cboxlength][c]{0.48} \\ 

 \hline 
 % & 1 & \colorbox{d2_r1_q1}{\makebox[\cboxlength][c]{1.9 $\pm$ 7.1}} & \colorbox{d2_r1_q2}{\makebox[\cboxlength][c]{0.8 $\pm$ 3.3}} & \colorbox{d2_r1_q3}{\makebox[\cboxlength][c]{0.3 $\pm$ 3.0}} \\ 
  & 1 & \colorbox{d2_r1_q1}{\makebox[\cboxlength][c]{1.7 $\pm$ 7.2}} & \colorbox{d2_r1_q2}{\makebox[\cboxlength][c]{0.6 $\pm$ 3.4}} & \colorbox{d2_r1_q3}{\makebox[\cboxlength][c]{0.3 $\pm$ 3.4}} \\ 
 & 2 & \colorbox{d2_r2_q1}{\makebox[\cboxlength][c]{1.0 $\pm$ 5.6}} & \colorbox{d2_r2_q2}{\makebox[\cboxlength][c]{-0.1 $\pm$ 3.6}} & \colorbox{d2_r2_q3}{\makebox[\cboxlength][c]{0.1 $\pm$ 3.6}} \\ 
 \multirow{2}{*}{\textbf{Patient}} & 3 & \colorbox{d2_r3_q1}{\makebox[\cboxlength][c]{2.3 $\pm$ 7.2}} & \colorbox{d2_r3_q2}{\makebox[\cboxlength][c]{2.0 $\pm$ 5.3}} & \colorbox{d2_r3_q3}{\makebox[\cboxlength][c]{2.2 $\pm$ 5.9}} \\ 
 \multirow{2}{*}{\textbf{questions}} & 4 & \colorbox{d2_r4_q1}{\makebox[\cboxlength][c]{1.9 $\pm$ 6.7}} & \colorbox{d2_r4_q2}{\makebox[\cboxlength][c]{0.0 $\pm$ 0.0}} & \colorbox{d2_r4_q3}{\makebox[\cboxlength][c]{0.0 $\pm$ 0.0}} \\ 
 & 5 & \colorbox{d2_r5_q1}{\makebox[\cboxlength][c]{0.9 $\pm$ 5.7}} & \colorbox{d2_r5_q2}{\makebox[\cboxlength][c]{0.4 $\pm$ 3.6}} & \colorbox{d2_r5_q3}{\makebox[\cboxlength][c]{0.4 $\pm$ 3.6}} \\ 
 & Pooled & \colorbox{d2_ra_q1}{\makebox[\cboxlength][c]{1.6 $\pm$ 6.5}}* & \colorbox{d2_ra_q2}{\makebox[\cboxlength][c]{0.6 $\pm$ 3.7}}* & \colorbox{d2_ra_q3}{\makebox[\cboxlength][c]{0.6 $\pm$ 3.9}}* \\ 
  & ICC & \makebox[\cboxlength][c]{0.67} & \makebox[\cboxlength][c]{0.31} & \makebox[\cboxlength][c]{0.21} \\ 

 \hline 
 & 1 & \colorbox{d3_r1_q1}{\makebox[\cboxlength][c]{3.4 $\pm$ 7.5}} & \colorbox{d3_r1_q2}{\makebox[\cboxlength][c]{0.5 $\pm$ 2.5}} & \colorbox{d3_r1_q3}{\makebox[\cboxlength][c]{0.1 $\pm$ 4.5}} \\ 
 & 2 & \colorbox{d3_r2_q1}{\makebox[\cboxlength][c]{2.3 $\pm$ 6.5}} & \colorbox{d3_r2_q2}{\makebox[\cboxlength][c]{0.6 $\pm$ 4.4}} & \colorbox{d3_r2_q3}{\makebox[\cboxlength][c]{0.4 $\pm$ 4.2}} \\ 
 \multirow{2}{*}{\textbf{Progress}} & 3 & \colorbox{d3_r3_q1}{\makebox[\cboxlength][c]{2.7 $\pm$ 6.3}} & \colorbox{d3_r3_q2}{\makebox[\cboxlength][c]{1.0 $\pm$ 4.4}} & \colorbox{d3_r3_q3}{\makebox[\cboxlength][c]{0.9 $\pm$ 3.7}} \\ 
 \multirow{2}{*}{\textbf{notes}} & 4 & \colorbox{d3_r4_q1}{\makebox[\cboxlength][c]{2.5 $\pm$ 7.2}} & \colorbox{d3_r4_q2}{\makebox[\cboxlength][c]{0.5 $\pm$ 6.8}} & \colorbox{d3_r4_q3}{\makebox[\cboxlength][c]{1.7 $\pm$ 6.9}} \\ 
 & 5 & \colorbox{d3_r5_q1}{\makebox[\cboxlength][c]{2.0 $\pm$ 6.8}} & \colorbox{d3_r5_q2}{\makebox[\cboxlength][c]{-0.8 $\pm$ 4.5}} & \colorbox{d3_r5_q3}{\makebox[\cboxlength][c]{-0.1 $\pm$ 1.2}} \\ 
 & Pooled & \colorbox{d3_ra_q1}{\makebox[\cboxlength][c]{2.6 $\pm$ 6.9}}* & \colorbox{d3_ra_q2}{\makebox[\cboxlength][c]{0.4 $\pm$ 4.8}} & \colorbox{d3_ra_q3}{\makebox[\cboxlength][c]{0.6 $\pm$ 4.5}}* \\ 
  & ICC & \makebox[\cboxlength][c]{0.77} & \makebox[\cboxlength][c]{0.74} & \makebox[\cboxlength][c]{0.42} \\ 

  \hline
  \multirow{2}{*}{\textbf{Overall}} & Pooled & \colorbox{d99_r99_q1}{\makebox[\cboxlength][c]{2.3 $\pm$ 5.8}}* & \colorbox{d99_r99_q2}{\makebox[\cboxlength][c]{0.8 $\pm$ 3.7}}* & \colorbox{d99_r99_q3}{\makebox[\cboxlength][c]{0.4 $\pm$ 4.0}}* \\
  & ICC & \makebox[\cboxlength][c]{0.63} & \makebox[\cboxlength][c]{0.56} & \makebox[\cboxlength][c]{0.38} \\
\hline
\end{tabular}}
\end{center}
\label{tab:reader-study-appendix}
\end{table*}
\begin{figure*}[t]
    \centering
    \includegraphics[width = 1.\textwidth]{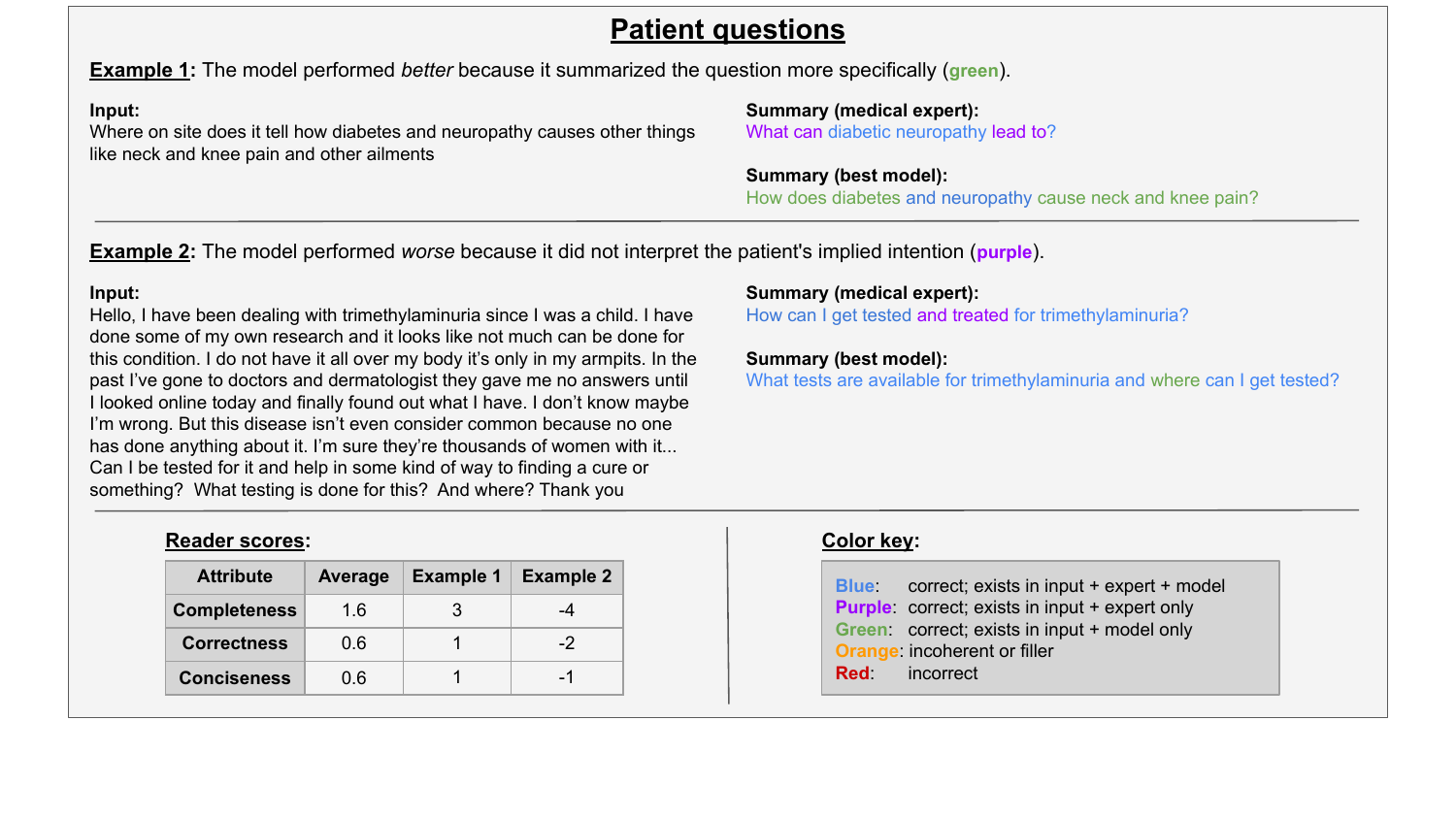}
    \caption{\textbf{Annotation: patient health questions.} The table (lower left) contains reader scores for these two examples and the task average across all samples.
    }
    \label{fig:clin-summ-example-chq}
\end{figure*}

\begin{figure*}[t]
    \centering
    \includegraphics[width = 1.\textwidth]{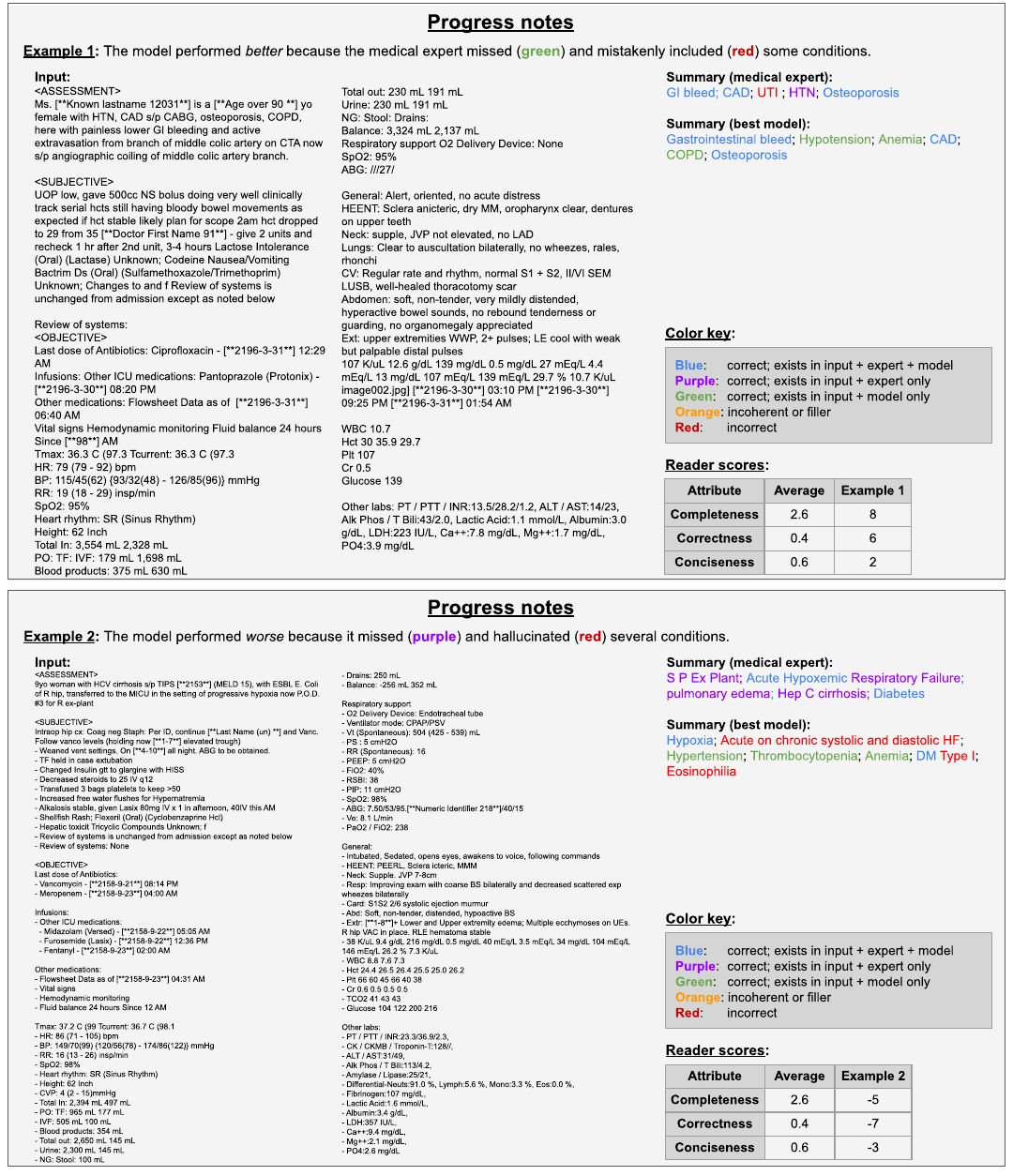}
    \caption{\textbf{Annotation: progress notes.} The tables (lower right) contain reader scores for this example and the task average across all samples.
    }
    \label{fig:clin-summ-example-pls}
\end{figure*}

\begin{figure*}[t]
    \centering
    \includegraphics[width = 1.\textwidth]{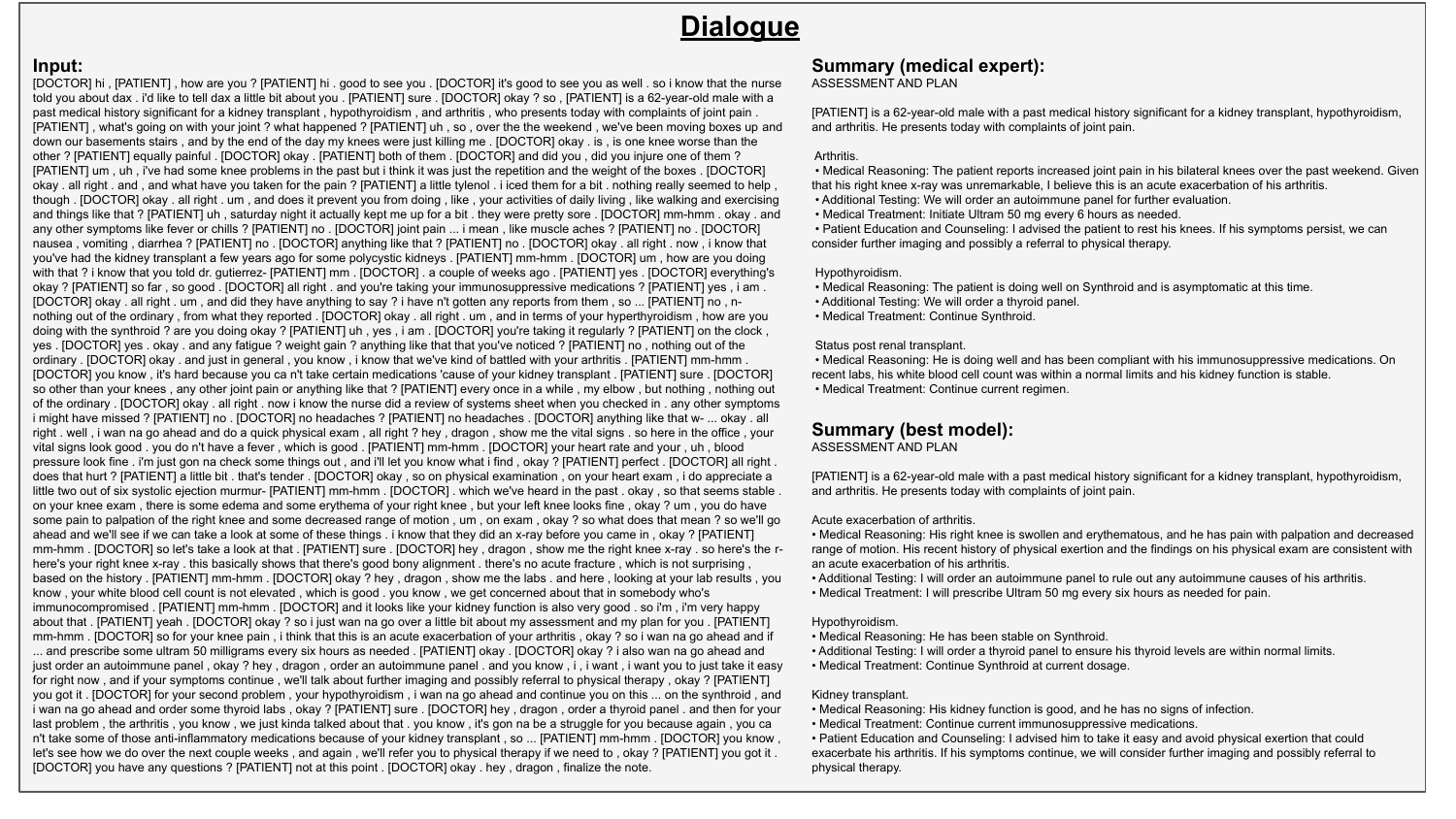}
    \caption{\textbf{Example results: doctor-patient dialogue.} Note this task is discluded from the reader study due to the unwieldiness of a reader parsing many transcribed conversations and lengthy text.
    }
    \label{fig:clin-summ-example-d2n}
\end{figure*}
\clearpage
\setlength{\tabcolsep}{6pt} % spacing b/w cols, default 6pt
\setlength{\extrarowheight}{1.5pt}

\begin{table*}[b]
  {\caption{
  \hhll{Comparison of our general approach (GPT-4 using ICL) against baselines specific to each individual dataset. We note the focal point of our study is not to achieve state-of-the-art quantitative results, especially given the discordance between NLP metrics and reader study scores. A - indicates the metric was not reported; a $^{\circ}$ indicates the dataset was preprocessed differently.}
  }}
\begin{center}

{\begin{tabular}{ll|cccc}
  \hline
  \textbf{Dataset} & \textbf{Baseline} & \textbf{BLEU} & \textbf{ROUGE-L} & \textbf{BERTScore} & \textbf{MEDCON} \\
  \hline
  \multirow{2}{*}{Open-i} & Ours & 46.0 & \textbf{68.2} & 94.7 & 64.9 \\
  & ImpressionGPT~\cite{ma2023impressiongpt} & - & 65.4 & - & - \\
   \hline
   \multirow{3}{*}{MIMIC-CXR} & Ours & \textbf{29.6} & \textbf{53.8} & \textbf{91.5} & 55.6 \\
   & RadAdapt~\cite{van2023radadapt} & 18.9 & 44.5 & 90.0 & - \\
   & ImpressionGPT~\cite{ma2023impressiongpt} & - & 47.9 & - & - \\
   \hline
   \multirow{3}{*}{MIMIC-III} & Ours & 11.5 & 34.5 & 89.0 & 36.5 \\
   & RadAdapt~\cite{van2023radadapt} & \textbf{16.2} & \textbf{38.7} & \textbf{90.2} & - \\
   & Med-PaLM M~\cite{tu2023towards} & 15.2 & 32.0 & - & - \\
   \hline
   \multirow{2}{*}{Patient questions} & Ours & 10.7 & 37.3 & 92.5 & 59.8 \\
  & ECL$^{\circ}$~\cite{wei2023medical} & - & \textbf{50.5} & - & - \\
  \hline
  \multirow{2}{*}{Progress notes} & Ours & 3.4 & 27.2 & 86.1 & 31.5 \\
  & CUED~\cite{manakul2023cued} & - & \textbf{30.1} & - & - \\
  \hline
  \multirow{2}{*}{Dialogue} & Ours & 26.9 & 42.9 & 90.2 & \textbf{59.9} \\
  & ACI-Bench$^{\circ}$~\cite{yim2023aci} & - & \textbf{45.6} & - & 57.8 \\
\hline
\end{tabular}}
\end{center}
\label{tab:baselines}
\end{table*}
\clearpage

\end{refsection}
\end{document}